\documentclass[lettersize,journal]{IEEEtran}
\usepackage{amsmath,amsfonts}
\usepackage{algorithmic}
\usepackage{algorithm}
\usepackage{array}
\usepackage[caption=false,font=normalsize,labelfont=sf,textfont=sf]{subfig}
\usepackage{textcomp}
\usepackage{stfloats}
\usepackage{url}
\usepackage{verbatim}
\usepackage{graphicx}
\usepackage{cite}
\usepackage{booktabs}   
\usepackage{tabularx}
\usepackage{multirow}  
\usepackage{xcolor}
\usepackage{makecell}   
\usepackage{graphicx}   
\usepackage{adjustbox}
\usepackage{bm}
\usepackage{quantikz}
\usepackage{placeins} 
\usepackage[colorlinks, linkcolor=blue, citecolor=blue, urlcolor=blue]{hyperref}
\usepackage[capitalize]{cleveref}
\usepackage{tikz}
\usetikzlibrary{positioning, arrows.meta, calc, shapes.geometric, fit}

\begin{document}
	
	\title{Quantum Incremental Learning \\ with Mixed State Prototypes}
	
	\author{Yu Wu, Qianli Zhou, Xinyang Deng, Wen Jiang, Kang Hao Cheong, Witold Pedrycz, \IEEEmembership{Life Fellow, IEEE}

	\thanks{Yu Wu, is with the School of Electronics and Information, Northwestern Polytechnical University, Xi’an, 710072, China and also with the School of Computer Science and Technology, Northwestern Polytechnical University, Xi’an, 710072, China.}

	\thanks{Qianli Zhou, Xinyang Deng, and Wen Jiang are with the School of Electronics and Information, Northwestern Polytechnical University, Xi’an, 710072, China.}
	
	\thanks{Kang Hao Cheong is with the School of Physical and Mathematical Sciences, Nanyang Technological University, Singapore 637371, and also with the AI College of Computing and Data Science, Nanyang Technological University, Singapore 639798 (e-mail: kanghao.cheong@ntu.edu.sg).}

	\thanks{Witold Pedrycz is with the Department of Electrical and Computer Engineering, University of Alberta, Edmonton, AB T6G 2R3, Canada, the Institute of Systems Engineering, Macau University of Science and Technology, Taipa 999078, Macau SAR, China, Systems Research Institute, Polish Academy of Sciences, 00-901 Warsaw, Poland, and Research Center of Performance and Productivity Analysis, Istinye University, Istanbul, Türkiye (e-mail: wpedrycz@ualberta.ca).}

	\thanks{This work is partially supported by the Chinese Postdoctoral Science Foundation‌ (Grant No. 2025M784414), and Postdoctoral Innovative Talent Support Program (Grant No. BX2026481).}

	\thanks{Manuscript received April 19, 2021; revised August 16, 2021.}}

	\markboth{Journal of \LaTeX\ Class Files,~Vol.~14, No.~8, August~2021}%
	{Shell \MakeLowercase{\textit{et al.}}: A Sample Article Using IEEEtran.cls for IEEE Journals}

	\maketitle
	
	\begin{abstract}
		Incremental learning models are required to learn new classes sequentially without catastrophic forgetting, while operating under parameter and memory constraints. In the Noisy Intermediate-Scale Quantum (NISQ) era, although quantum neural networks offer advantages in feature mapping, hardware limitations restrict circuit width. Furthermore, traditional quantum classifiers are constrained by the number of orthogonal basis states, limiting their capacity to accommodate a continually growing number of categories. Thus, we introduce a novel quantum incremental learning framework based on trainable mixed-state prototypes. Its original design incorporates new classes by adding class prototypes rather than increasing the circuit width of the shared quantum backbone. The use of mixed-state prototypes is another key contribution, since they have representation capabilities to represent information than a single pure-state prototype. And the decomposable mixed-state calculation provides lower production costs and a convenient Hilbert–Schmidt (HS) distance metric for classification. Simulation results show that our model achieves high-dimensional feature concentration using a minimal number of qubits, while demonstrating lower computational complexity and robust representation in incremental learning tasks compared with classical baselines.
	\end{abstract}
	
	
	\section{Introduction}
	\label{sec:Introduction}

\IEEEPARstart{D}{eep} learning has achieved remarkable success under the assumption that all training data is available at once. However, real-world data typically arrives in a continuous, ever-changing stream. Iteratively retraining a model from scratch to integrate new classes is computationally exorbitant and inefficient for resource-constrained applications \cite{zhou2024class}. Consequently, incremental learning has emerged to continuously update models with new knowledge while minimizing storage and computing overhead \cite{zhang2025few, masana2022class}.

In incremental learning, the fundamental challenge is catastrophic forgetting, where neural networks erase previously learned representations upon adapting to new tasks \cite{gao2022r, mccloskey1989catastrophic}. Early works introduced memory replay to explicitly rehearse a small buffer of historical exemplars \cite{rebuffi2017icarl,wang2022foster,douillard2020podnet}, alongside parameter regularization and knowledge distillation to constrain critical weight updates\cite{kirkpatrick2017overcoming,li2017learning,smith2021always}. As storing raw data is often restricted by memory limits or privacy concerns, research subsequently explored exemplar-free statistical prototype matching \cite{huang2026incomplete,huang2026dual}, extracting and freezing abstract representations to avoid iterative retraining\cite{petit2023fetril,goswami2023fecam,zhu2021prototype}. More recently, dynamic networks have been proposed to mitigate forgetting by expanding architectural branches\cite{van2022three,zhao2026multidomain}. Despite these diverse efforts, classical incremental learning still confronts the bottleneck of computational overhead and residual representational drift, which limits their real-world scalability \cite{zhou2024class}.

Harnessing quantum mechanics, quantum computing introduces a fundamentally distinct paradigm for information processing, catalyzing the rapid exploration of quantum machine learning\cite{xu2024quantum,zhan2026ternary}. In the NISQ era, hybrid architectures based on variational quantum circuits (VQC) have shown immense promise\cite{li2025efficient}. By projecting classical data into an exponentially expansive Hilbert space, VQCs facilitate the efficient manipulation of highly non-linear features \cite{shi2025quantum, cerezo2021variational, xiao2026adaptive}. Thus, quantum models achieve competitive accuracy with significantly reduced parameters compared to classical networks \cite{shindi2023model}. Moreover, quantum entanglement captures non-local correlations, providing a physical foundation for enhancing model expressiveness \cite{deng2023novel, xiao2022negation}. Recent studies have extended quantum neural architectures to tasks including human reliability analysis\cite{su2026dependence}, multimodal feature fusion \cite{wu2026feature,yang2025multimodal} and protein-folding prediction \cite{shi2026qsyncfold}. Furthermore, Shi et al. \cite{shi2024qsan} creatively proposed a quantum self-attention mechanism based on quantum logic similarity, which can efficiently and accurately calculate attention scores and outputs while reducing intermediate measurements, further exploring the development potential of quantum machine intelligence in the NISQ era.

Despite these theoretical merits, classical incremental learning strategies cannot be directly quantized, as the structural differences between quantum and classical networks\cite{jiang2022quantum}. In classical deep learning, accommodating new classes is often as straightforward as expanding the final linear layer. However, a large class of commonly used quantum classifiers constructs class scores from a fixed set of basis-state probabilities or expectation values \cite{du2023problem, nghiem2021unified}. As the label space grows, these classifiers may require an expanded measurement or readout rule, additional observables, or more class-dependent parameters. In basis-state encoding schemes, accommodating additional classes may also require additional qubits. Crucially, unlike modular classical expansion, appending a qubit fundamentally alters the global quantum state and the underlying Hilbert space. This structural shift disrupts previously learned entanglement correlations, exacerbating catastrophic forgetting \cite{zhang2026experimental}. Furthermore, blindly increasing the circuit width or depth inevitably triggers the barren plateau phenomenon, rendering the quantum network untrainable \cite{yuan2023optimal}.

To circumvent this structural expansion dilemma, quantum metric learning emerges as a promising paradigm. By classifying instances based on state distances rather than relying on explicit parameterized classification heads, it naturally bypasses the need for circuit widening. Its effectiveness heavily depends on exploring advanced quantum information representations \cite{lloyd2020quantum,huang2026individual}. In quantum feature mapping, pure states encode individual inputs into single unit state vectors in the Hilbert space\cite{dong2019learning}. Consequently, they lack the statistical capacity to represent a diverse distribution compared to mixed states. By formulating information as a convex combination of orthogonal states, mixed states encapsulate data randomness and complex intra-class variance \cite{ezzell2023quantum}. Compared to classical information representations, this quantum method exponentially broadens the representation dimensionality. Furthermore, it provides physically grounded distance metrics, such as the HS distance, trace distance, and Bures distance\cite{travnivcek2019experimental}. Ultimately, these comprehensive advantages help to solve the problems in incremental learning tasks. \cite{asadi2023prototype}.

	\begin{figure*}[!t]
	\centering
	\begin{tikzpicture}[
		font=\small,
		>=stealth,
		box/.style={
			draw=black!80,
			thick,
			rounded corners=3pt,
			align=center,
			minimum width=1.55cm,
			minimum height=0.8cm
		},
		modelbox/.style={
			draw=black!80,
			thick,
			rounded corners=3pt,
			align=center,
			minimum width=1.6cm,
			minimum height=1.8cm
		},
		arrow/.style={->, thick, draw=black!70},
		darrow/.style={<->, thick, dashed, draw=red!70},
		panel/.style={
			draw=gray!40,
			densely dotted,
			rounded corners=5pt,
			minimum width=5.4cm,
			minimum height=4.0cm
		},
		title/.style={font=\bfseries, text=black!90}
		]
		
		\node[panel] (pA) at (0,0) {};
		\node[title] at ($(pA.north)+(0,-0.35)$) {(a) Exemplar Replay};
		
		\node[box, fill=blue!8] (a_data) at ($(pA.center)+(-1.4,0.75)$)
		{New Data\\$\mathcal{D}_t$};
		
		\node[box, fill=orange!10] (a_mem) at ($(pA.center)+(-1.4,-0.85)$)
		{Memory\\$\mathcal{E}_{t-1}$};
		
		\node[box, fill=orange!10, dashed] (a_old) at ($(pA.center)+(1.0,0.75)$)
		{Old Model\\$\Theta_{t-1}$};
		
		\node[box, fill=green!10] (a_new) at ($(pA.center)+(1.0,-0.85)$)
		{New Model\\$\Theta_t$};
		
		\draw[arrow] (a_old.south) -- coordinate(a_finetune) (a_new.north);
		
		\node[right=2pt, font=\scriptsize] at (a_finetune) {Finetune};
		
		\coordinate (a_bus_top) at ($(a_data.east)+(0.45,0)$);
		\coordinate (a_bus_bottom) at ($(a_mem.east)+(0.45,0)$);
		\coordinate (a_bus_mid) at ($(a_bus_top)!0.5!(a_bus_bottom)$);
		
		\draw[thick, draw=black!70] (a_bus_top) -- (a_bus_bottom);
		
		\draw[thick, draw=black!70] (a_data.east) -- (a_bus_top);
		\draw[thick, draw=black!70] (a_mem.east) -- (a_bus_bottom);
		
		\draw[arrow] (a_bus_mid) -- (a_finetune);
		
		\draw[arrow, densely dashed, draw=black!60]
		(a_data.south) -- node[left, font=\scriptsize] {Update} (a_mem.north);
		
		\draw[arrow, dashed, draw=black!60]
		(a_new.east) -- ++(0.35,0) |- node[pos=0.55, right, font=\scriptsize] {} (a_old.east);
		
		\node[panel] (pB) at (5.8,0) {};
		\node[title] at ($(pB.north)+(0,-0.35)$) {(b) Regularization};
		
		\node[box, fill=blue!8] (b_data) at ($(pB.center)+(-1.5,-0.85)$)
		{New Data\\$\mathcal{D}_t$};
		
		\node[box, fill=orange!10, dashed] (b_old) at ($(pB.center)+(0.95,0.75)$)
		{Old Model\\$\Theta_{t-1}$\\{\scriptsize (Frozen)}};
		
		\node[box, fill=green!10] (b_new) at ($(pB.center)+(0.95,-0.85)$)
		{New Model\\$\Theta_t$};
		
		\draw[arrow] (b_data.east) -- (b_new.west);
		
		\draw[arrow, dashed, draw=black!80] ([xshift=-0.15cm]b_new.north) -- ([xshift=-0.15cm]b_old.south);
		\draw[arrow, draw=black!80] ([xshift=0.15cm]b_old.south) -- ([xshift=0.15cm]b_new.north);
		
		\node[right, font=\scriptsize, align=left, text=black!90] 
		at ($(b_old.south)!0.5!(b_new.north) + (0.2,0)$) 
		{Regularize};

		\node[panel] (pC) at (11.6,0) {};
		\node[title] at ($(pC.north)+(0,-0.35)$) {(c) Prototype Classification};
		
		\node (c_input) at ($(pC.center)+(-2.25,0)$) {$\mathbf{x}$};
		
		\node[modelbox, fill=green!10, minimum height=1.5cm] (c_enc)
		at ($(pC.center)+(-1.1,0)$)
		{Encoder\\$F$};
		
		\draw[arrow] (c_input.east) -- (c_enc.west);
		
		\draw[
		thick,
		rounded corners=4pt,
		fill=gray!6,
		draw=gray!60
		] ($(pC.center)+(0.2,-1.2)$) rectangle ($(pC.center)+(2.55,1.2)$);

		\begin{scope}[shift={(pC.center)}]
			
			\filldraw[
			fill=blue!8,
			draw=blue!80!black,
			thick,
			rotate around={28:(1.30,0.45)}
			] (1.00,0.45) ellipse (0.85 and 0.34);
			
			\filldraw[
			fill=green!8,
			draw=green!55!black,
			thick,
			rotate around={-25:(1.95,-0.45)}
			] (1.95,-0.45) ellipse (0.63 and 0.29);
			
			\foreach \x/\y in {
				0.42/0.05, 0.56/0.23, 0.64/0.45, 0.70/0.13, 0.78/0.33, 0.88/0.55, 0.94/0.21, 1.02/0.39, 1.10/0.09, 1.16/0.51, 1.22/0.27, 1.28/0.61, 1.32/0.17, 1.38/0.39, 1.14/-0.05, 0.82/0.01, 0.96/0.63, 1.24/0.41
			}{
				\node[
				cross out,
				draw=blue!50!white,
				line width=0.35pt,
				minimum size=1.5pt,
				inner sep=0pt
				] at (\x,\y) {};
			}
			
			\node[
			cross out,
			draw=blue!90!black,
			line width=0.4pt,
			minimum size=1.5pt,
			inner sep=0.5pt
			] (c_blue_center) at (1.00,0.3) {};
			
			\foreach \x/\y in {
				1.62/-0.18, 1.72/-0.30, 1.78/-0.50, 1.86/-0.20,
				1.92/-0.40, 1.98/-0.56, 2.04/-0.30, 2.08/-0.48,
				2.14/-0.64, 2.18/-0.40, 2.20/-0.24, 1.74/-0.64,
				1.88/-0.58, 2.00/-0.16, 2.12/-0.44, 1.66/-0.44
			}{
				\node[
				regular polygon,
				regular polygon sides=3,
				draw=green!75!black,
				fill=white,
				minimum size=1pt,
				inner sep=0.5pt
				] at (\x,\y) {};
			}
			
			\node[
			regular polygon,
			regular polygon sides=3,
			draw=green!50!black,
			fill=white,
			thick,
			minimum size=2pt,
			inner sep=0.5pt
			] (c_q) at (1.95,-0.45) {};

			\filldraw[
			fill=orange!8,
			draw=orange!80!black,
			thick,
			rotate around={18:(0.85,-0.65)}
			] (0.85,-0.65) ellipse (0.58 and 0.26);
			
			\foreach \x/\y in {
				0.48/-0.78, 0.58/-0.62, 0.66/-0.86, 0.72/-0.52,
				0.80/-0.72, 0.88/-0.58, 0.95/-0.84, 1.02/-0.66,
				1.10/-0.50, 1.18/-0.76, 0.62/-0.72, 1.00/-0.92
			}{
				\node[
				circle,
				draw=orange!80!black,
				fill=white,
				minimum size=1pt,
				inner sep=0.5pt
				] at (\x,\y) {};
			}
			
			\node[
			circle,
			draw=orange!90!black,
			fill=white,
			thick,
			minimum size=2pt,
			inner sep=0.5pt
			] (c_orange_center) at (0.85,-0.65) {};
			
		\end{scope}
		
		\draw[arrow, dashed] (c_enc.east) -- ($(pC.center)+(0.7,-0.55)$);
		
		\node[font=\scriptsize, text=gray!80!black]
		at ($(pC.center)+(1.38,0.97)$)
		{Embedding Space};
		
	\end{tikzpicture}
	
	\caption{Illustration of incremental learning strategies. 
		(a) Exemplar replay; (b) Regularization; (c) Prototype classification.}
	\label{fig:IL strategies}
\end{figure*}
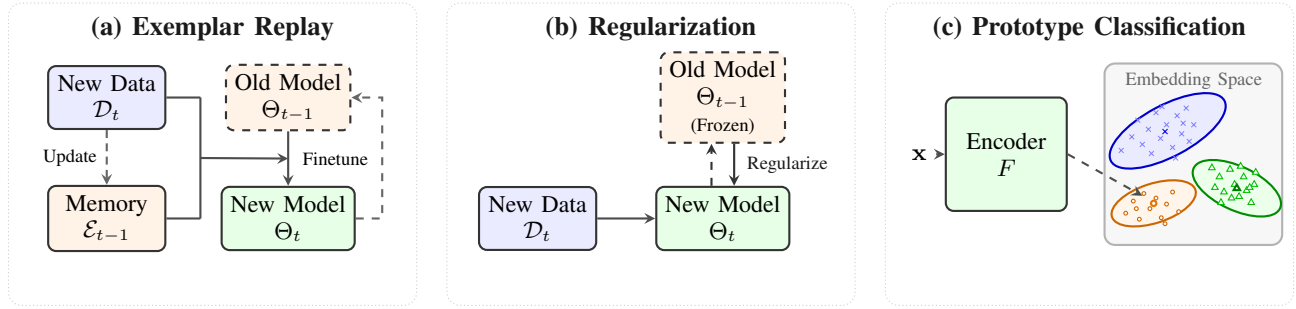

Motivated by the representational power of mixed states and the physical constraints of quantum circuits, this paper proposes a novel quantum incremental learning framework based on mixed-state prototypes. To transcend the capacity bottleneck of traditional quantum classifiers, our design shifts from explicit softmax regression classification to an implicit, distance-based prototype matching mechanism. Moreover, recognizing the difficulty of dynamically altering quantum network, we design a hybrid architecture: we append a lightweight, expandable classical module (i.e., an MLP) to manage the incremental class adjustments without disrupting the underlying quantum states. The main contributions of this work are summarized as follows:(1) A quantum prototype framework with an expandable module that avoids the capacity constraint associated with orthogonal basis states. (2) Mixed-state prototypes via dynamic weighting of pure states avoid redundant qubits and provide PCA-like noise-filtering properties. (3) Experiments on static and incremental classification tasks demonstrate high-dimensional semantic concentration, exhibiting parameter advantages and noise robustness.
	
	The rest of this paper is organized as follows. \cref{sec:preliminaries} introduces the preliminaries of incremental learning and mixed-state compiling. \cref{sec:Mixed-state based quantum prototype classifier} details the construction of the mixed-state quantum prototype classifier. \cref{sec:Quantum Incremental Learning Framework} presents the overall incremental learning framework at length. \cref{sec:Experiment} provides the experimental evaluation, and \cref{sec:conclusion} concludes the paper  and discusses some directions.

	\section{Preliminaries}
	\label{sec:preliminaries}
	This section provides a concise overview of the key concepts in incremental learning and quantum mixed state compilation that form the foundation of our proposed framework.

	\subsection{Incremental learning strategies}
	\label{subsec:Incremental learning strategies}
	
	Incremental learning requires a model to sequentially learn new classes from a continuous data stream without catastrophically forgetting previously acquired knowledge. To strike a balance between the plasticity required to learn new tasks and the stability needed to retain old ones, current classical incremental learning frameworks generally rely on three fundamental strategies, as \cref{fig:IL strategies} shows.

	\subsubsection{Exemplar replay strategy} 
	Replay mechanisms mitigate forgetting by preserving a bounded memory of old data. A prominent example is iCaRL \cite{rebuffi2017icarl}, which utilizes a herding algorithm to select and store a small set of highly representative samples, for each observed class. During the incremental training phase, these exemplars are replayed alongside new class instances. This allows the model to retrospectively review past knowledge, effectively preventing the decision boundaries from catastrophic forgetting.
	
	\subsubsection{Regularization techniques} 
	Instead of directly reusing raw data, regularization methods impose mathematical constraints on the network's optimization trajectory. Parameter-based regularization approaches like EWC \cite{kirkpatrick2017overcoming} evaluate the importance of individual network weights for past tasks and penalize drastic deviations in these critical parameters. Furthermore, knowledge distillation techniques such as LwF \cite{li2017learning} mitigate the representational drift of the network. They force the updated model to mimic the output logits or intermediate feature representations of the frozen old model, thereby implicitly preserving the topological structure of past knowledge.
	
	\subsubsection{Prototype-based classification} 
	Due to the data imbalance between old and new classes, traditional classifiers suffer from task-recency bias in incremental learning. To address this, prototype-based inference relies on distance metrics within the embedding space\cite{asadi2023prototype}. Methods like the nearest-mean-of-exemplars classifier in iCaRL or centroid-based classifiers in FeTrIL and FeCAM \cite{petit2023fetril, goswami2023fecam} represent each class by a global prototype. An input instance is classified by assigning it to the nearest class prototype. This decoupling of feature extraction from explicit classification effectively alleviates the inherent prediction bias and serves as a theoretical cornerstone for our proposed quantum prototype design.

	\subsection{Quantum mixed state compiling}
	\label{subsec:Quantum mixed state compiling}
	
	Quantum state compilation aims to learn a parameterized quantum circuit that can prepare a reliable approximation of target state. While early works primarily focused on pure states, compiling mixed states provides a more generalized framework for handling classical probabilities, noise, and complex data distributions in the NISQ era\cite{cerezo2020variational}.
	
	\subsubsection{Density matrix and mixed state representation} 
	In quantum mechanics, a closed quantum system with complete information is described by a pure state, denoted by a state vector $|\psi\rangle$ in a Hilbert space. However, in practical open quantum systems, particularly within the NISQ era, a principal system inevitably interacts with its environment and undergoes decoherence. Tracing out the environment produces a reduced mixed state. A mixed state is described with a density matrix $\rho$. It is a positive-semidefinite Hermitian operator with a unit trace $\text{Tr}(\rho) = 1$. According to the spectral decomposition theorem, any mixed state $\rho$ can be expressed as a probabilistic ensemble of orthogonal pure states:
	\begin{equation}
		\rho = \sum_{i} \lambda_i |v_i\rangle \langle v_i|,
		\label{eq:density_matrix}
	\end{equation}
	where $\lambda_i \geq 0$ are the eigenvalues representing the classical probability of the system being in the corresponding pure eigenstate $|v_i\rangle$, satisfying $\sum_i \lambda_i = 1$.

	\begin{figure}[htbp]
		\centering
		\resizebox{\linewidth}{!}{
			\begin{tikzpicture}[
				>={Latex[length=2.5mm, width=1.8mm]},
				operator/.style={circle, draw=black, thick, inner sep=0pt, minimum size=0.55cm, fill=white}
				]
				
				\newcommand{\drawthreeD}[8]{
					\filldraw[fill=#6, draw=black, thick] (#1, #2) rectangle (#1+#3, #2+#4); 
					\filldraw[fill=#7, draw=black, thick] (#1, #2+#4) -- (#1+#5, #2+#4+#5) -- (#1+#3+#5, #2+#4+#5) -- (#1+#3, #2+#4) -- cycle; 
					\filldraw[fill=#8, draw=black, thick] (#1+#3, #2) -- (#1+#3+#5, #2+#5) -- (#1+#3+#5, #2+#4+#5) -- (#1+#3, #2+#4) -- cycle; 
				}
				
				\drawthreeD{-1.0}{-4.0}{0.6}{4.4}{0.15}{black}{black!80}{black!60}
				\node[font=\normalsize, text=white] at (-0.7, -1.8) {$\bm{\rho}_{\text{tar}}$};
				
				\node (in1) at (0.3, 0.4) {$|1\rangle$};
				\drawthreeD{0.8}{0.0}{1.6}{0.8}{0.1}{blue!5}{blue!10}{blue!20}
				\node[font=\normalsize] at (1.6, 0.4) {VQC};
				\drawthreeD{3.8}{0.0}{1.6}{0.8}{0.1}{white}{black!5}{black!10}
				\node[font=\normalsize] at (4.6, 0.4) {Swap Test};
				
				\draw[->, thick] (in1.east) -- (0.8, 0.4);
				\draw[->, thick] (2.45, 0.45) -- (3.8, 0.45) node[pos=0.35, above] {$|\phi_1\rangle$};
				
				\node (in2) at (0.3, -1.4) {$|2\rangle$};
				\drawthreeD{0.8}{-1.8}{1.6}{0.8}{0.1}{blue!5}{blue!10}{blue!20}
				\node[font=\normalsize] at (1.6, -1.4) {VQC};
				\drawthreeD{3.8}{-1.8}{1.6}{0.8}{0.1}{white}{black!5}{black!10}
				\node[font=\normalsize] at (4.6, -1.4) {Swap Test};
				
				\draw[->, thick] (in2.east) -- (0.8, -1.4);
				\draw[->, thick] (2.45, -1.35) -- (3.8, -1.35) node[pos=0.35, above] {$|\phi_2\rangle$};
				
				\node (inK) at (0.3, -3.6) {$|K\rangle$};
				\drawthreeD{0.8}{-4.0}{1.6}{0.8}{0.1}{blue!5}{blue!10}{blue!20}
				\node[font=\normalsize] at (1.6, -3.6) {VQC};
				\drawthreeD{3.8}{-4.0}{1.6}{0.8}{0.1}{white}{black!5}{black!10}
				\node[font=\normalsize] at (4.6, -3.6) {Swap Test};
				
				\draw[->, thick] (inK.east) -- (0.8, -3.6);
				\draw[->, thick] (2.45, -3.55) -- (3.8, -3.55) node[pos=0.35, above] {$|\phi_K\rangle$};
				
				\draw[thick] (-0.2, -2) -- (3.3, -2);
				\draw[thick] (3.3, 0.2) -- (3.3, -3.8); 
				\draw[->, thick] (3.3, 0.2) -- (3.8, 0.2);   
				\draw[->, thick] (3.3, -1.6) -- (3.8, -1.6); 
				\draw[->, thick] (3.3, -3.8) -- (3.8, -3.8); 
				
				
				\node[operator] (ot1) at (6.8, 0.45) {};
				\draw[thick] (ot1.north west)--(ot1.south east); \draw[thick] (ot1.north east)--(ot1.south west);
				\node[anchor=south west, font=\normalsize, inner sep=1pt] at (ot1.north east) {$p_1$};
				
				\node[operator] (ot2) at (6.8, -1.35) {};
				\draw[thick] (ot2.north west)--(ot2.south east); \draw[thick] (ot2.north east)--(ot2.south west);
				\node[anchor=south west, font=\normalsize, inner sep=1pt] at (ot2.north east) {$p_2$};
				
				\node[operator] (otK) at (6.8, -3.55) {}; 
				\draw[thick] (otK.north west)--(otK.south east); \draw[thick] (otK.north east)--(otK.south west);
				\node[anchor=south west, font=\normalsize, inner sep=1pt] at (otK.north east) {$p_K$};
				
				\draw[->, thick] (5.45, 0.45) -- (ot1.west) node[midway, above] {$F_1$};
				\draw[->, thick] (5.45, -1.35) -- (ot2.west) node[midway, above] {$F_2$};
				\draw[->, thick] (5.45, -3.55) -- (otK.west) node[midway, above] {$F_K$};
				
				\node[operator, minimum size=0.7cm] (sum) at (8.8, -1.5) {};
				\draw[thick] (sum.north)--(sum.south); \draw[thick] (sum.west)--(sum.east);
				
				\draw[thick] (8.0, 0.45) -- (8.0, -3.55);
				\draw[-, thick] (ot1.east) -- (8.0, 0.45);
				\draw[-, thick] (ot2.east) -- (8.0, -1.35);
				\draw[-, thick] (otK.east) -- (8.0, -3.55);
				\draw[->, thick] (8.0, -1.5) -- (sum.west);
				
				\draw[->, thick] (sum.east) -- (9.8, -1.5);
				\drawthreeD{9.8}{-2.5}{0.5}{2.0}{0.1}{black}{black!80}{black!60}
				\node[font=\normalsize, text=white] at (10.05, -1.5) {$\hat{F}$};
				
				\node at (0.3, -2.4) {$\vdots$};
				\node at (1.6, -2.4) {$\vdots$};
				\node at (4.6, -2.4) {$\vdots$};
				\node at (8.0, -2.4) {$\vdots$}; 
				
			\end{tikzpicture}
		}
		\caption{HS distance measurement via CCPS. VQC and SWAP test is shared.}
		\label{fig:ccps_ansatz_workflow}
	\end{figure}
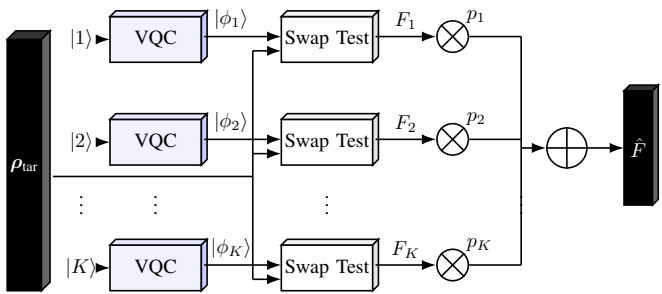

	\subsubsection{Convex Combination of Pure States ansatz}
	State-purification ansatzes prepare a larger pure state and trace out ancillary registers, which increases the required circuit width. The convex combination of pure states (CCPS) ansatz avoids this overhead by representing a rank-$R$ mixed state as a probabilistic mixture of transformed basis states \cite{ezzell2023quantum}:
	\begin{equation}
		\sigma_{\text{CCPS}}(\boldsymbol{\alpha}, R)
		:= \sum_{i=0}^{R-1} p_\phi(i) U_\theta |i\rangle
		\langle i| U_\theta^\dagger,
		\label{eq:ccps_ansatz}
	\end{equation}
	where $U_\theta$ is a parameterized unitary circuit, $|i\rangle$ denotes a selected computational basis state, and $p_\phi(i)$ is a trainable probability satisfying $\sum_i p_\phi(i)=1$. The trainable variables $\boldsymbol{\alpha}$ include both the circuit parameters $\theta$ and the probability parameters $\phi$.
	
	Under HS-distance optimization, a rank-$R$ CCPS state provides a variational low-rank approximation of the target mixed state, yielding a PCA-like procedure that retains its leading components \cite{ezzell2023quantum}. It controls the prototype rank without introducing ancillary qubits and can suppress low-contribution components. These properties make CCPS suitable for constructing the class prototypes in our framework.
	
	\subsubsection{Distance measurement via HS distance and SWAP test}
	\label{subsubsec:Distance measurement via HS distance and SWAP test}
	The HS distance is used to measure the distance between two quantum states, and also to describe the quantum distance in quantum geometry\cite{travnivcek2019experimental}. Using the Frobenius norm, the squared HS distance is $\mathcal{L} = \|\rho - \sigma\|_F^2$, which can be decomposed into state purities and their quantum overlap:
	\begin{equation}
		\mathcal{L} = \text{Tr}[\rho^2] + \text{Tr}[\sigma^2] - 2\text{Tr}[\rho \sigma].
		\label{eq:hs_expansion}
	\end{equation}
	CCPS's  method that can decouple the calculation into classical probability evaluations and a series of quantum SWAP tests between mixed and pure states.
	
	Firstly, due to the orthogonality of the computational basis states $|i\rangle$, the purity of the CCPS state simplifies entirely to the collision probability of its classical weights:
	\begin{equation}
		\text{Tr}[\sigma_{\text{CCPS}}^2] = \sum_{i=0}^{R-1} p_\phi(i)^2.
		\label{eq:ccps_purity}
	\end{equation}
	
\begin{figure}[!htbp]
		\centering
		\begin{adjustbox}{max width=\columnwidth}
			\begin{quantikz}[row sep=0.45cm, column sep=0.65cm]
				\lstick{$|0\rangle_a$} & \gate{H} & \ctrl{1} & \ctrl{2} & \gate{H} & \meter{} \\
				\lstick{$\rho_1$}      & \qw      & \swap{2} & \qw      & \qw      & \qw \\
				\lstick{$\rho_2$}      & \qw      & \qw      & \swap{2} & \qw      & \qw \\
				\lstick{$\phi_1$}      & \qw      & \targX{} & \qw      & \qw      & \qw \\
				\lstick{$\phi_2$}      & \qw      & \qw      & \targX{} & \qw      & \qw
			\end{quantikz}
		\end{adjustbox}
		\caption{SWAP test circuit for 2-qubit states $\rho$ and $\phi$.}
		\label{fig:swap_test_2qubit}
	\end{figure}
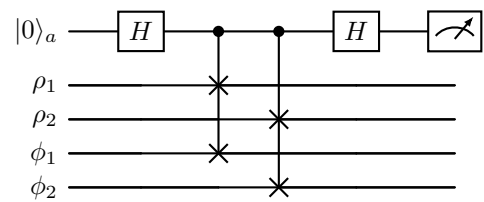

	Secondly, to expand the cross term $\text{Tr}[\rho \sigma_{\text{CCPS}}]$, let $|\psi_i(\theta)\rangle = U_\theta |i\rangle$ denote the $i$-th parameterized pure state generated by the ansatz. The cross term can be rewritten as:
	\begin{equation}
		\text{Tr}[\rho \sigma_{\text{CCPS}}] = \sum_{i=0}^{R-1} p_\phi(i) \text{Tr}\left[\rho |\psi_i(\theta)\rangle \langle \psi_i(\theta)|\right].
		\label{eq:ccps_crossterm}
	\end{equation}
	The term $\mathcal{F}_i(\theta) \equiv \text{Tr}\left[\rho |\psi_i(\theta)\rangle \langle \psi_i(\theta)|\right]$ represents the quantum state overlap, between the target mixed state $\rho$ and the pure state $|\psi_i(\theta)\rangle$. And each $\mathcal{F}_i(\theta)$ can be evaluated on quantum hardware using a SWAP test. The circuit is shown in \cref{fig:swap_test_2qubit}.
	
	Specifically, the SWAP test requires one $|0\rangle$ initialized ancillary qubit, utilizing a controlled-swap gate sandwiched between two Hadamard gates. By measuring the ancilla, the probability of observing the state $|0\rangle$, denoted as $P(0)$, is represented as:
	\begin{equation}
		P(0) = \frac{1}{2} + \frac{1}{2} \text{Tr}\left[\rho |\psi_i(\theta)\rangle \langle \psi_i(\theta)|\right].
		\label{eq:swap_test_prob}
	\end{equation}
	Thus, the required quantum state overlap can be directly expressed from the measurement statistic as $\mathcal{F}_i(\theta) = 2P(0) - 1$.
	
	By substituting Eq. (\ref{eq:ccps_purity}) and Eq. (\ref{eq:ccps_crossterm}) back into Eq. (\ref{eq:hs_expansion}), the overall HS distance is expressed as a combination of these single-parameter squares and state SWAP test results:
	\begin{equation}
		\mathcal{L}_{\text{CCPS}}(\boldsymbol{\theta}, \boldsymbol{\phi}) = \text{Tr}[\rho^2] + \sum_{i=0}^{R-1} p_\phi(i)^2 - 2 \sum_{i=0}^{R-1} p_\phi(i) \mathcal{F}_i(\theta).
		\label{eq:final_loss}
	\end{equation}
	
	During the CCPS's optimization, the target mixed state $\rho$ is fixed. Therefore, the loss function is reduced to:
	\begin{equation}
		\mathcal{L}_{\text{opt}}(\boldsymbol{\theta}, \boldsymbol{\phi}) = \sum_{i=0}^{R-1} p_\phi(i)^2 - 2 \sum_{i=0}^{R-1} p_\phi(i) \mathcal{F}_i(\theta).
		\label{eq:operational_loss}
	\end{equation}
	
	This decomposition is the main reason for using the HS distance in our framework. In dense-matrix simulation, evaluating the HS distance between two $d\times d$ states requires $O(d^2)$ operations. The trace distance, fidelity, Bures distance, and quantum Jensen--Shannon divergence generally require $O(d^3)$ operations because they involve spectral decompositions or matrix square roots. Therefore, the HS distance is more suitable for repeated prototype fitting and comparison.

	\begin{figure*}[htbp]
		\centering
		\includegraphics[trim={0.6cm 2.95cm 5.3cm 0.6cm}, clip, width=0.98\linewidth]{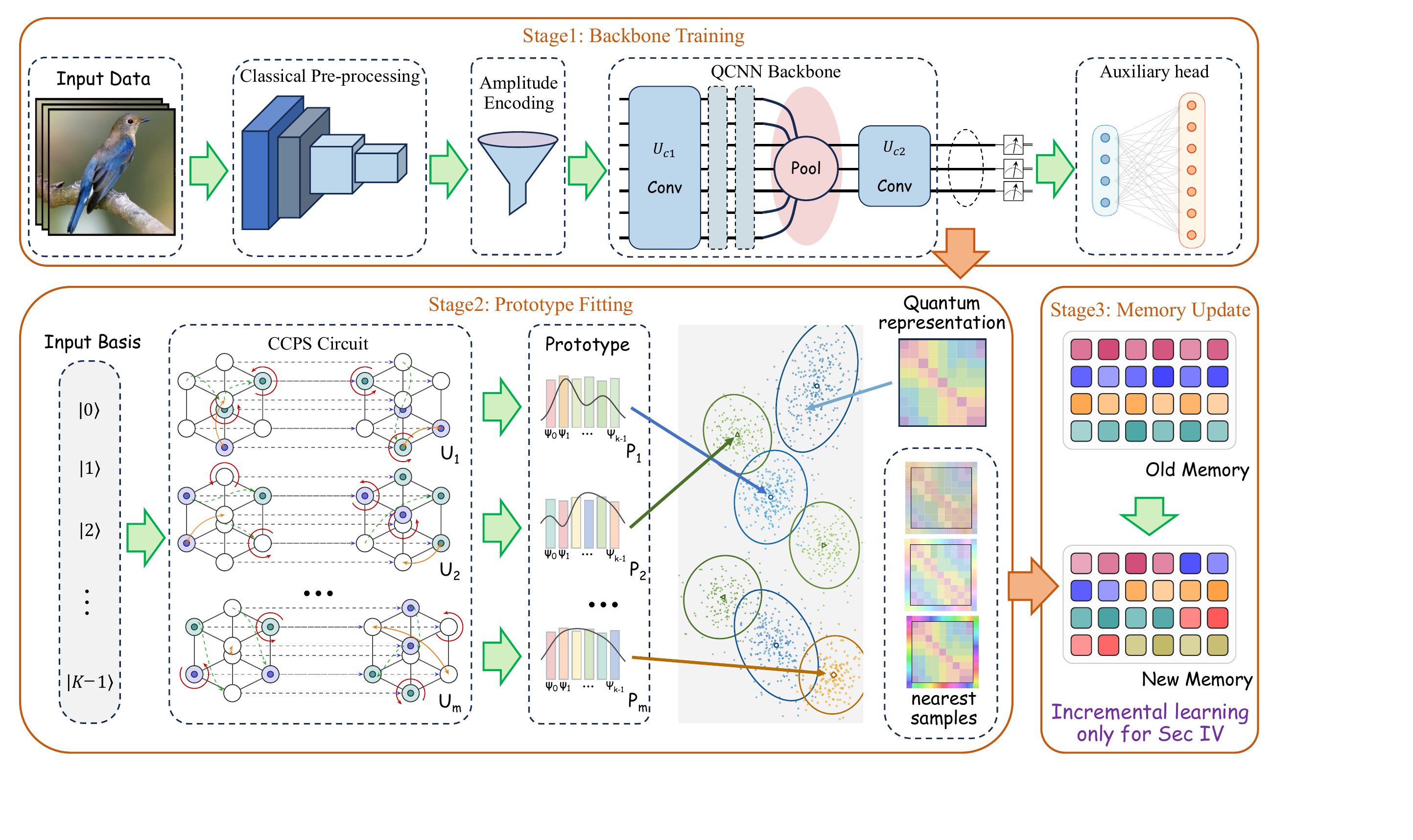}
		\caption[Overview of the proposed framework.]{Overview of the proposed framework. \textbf{Stage 1: Backbone training.} Classical inputs are compressed and amplitude-encoded into a QCNN. The backbone is optimized via an auxiliary head. \textbf{Stage 2: Prototype Fitting.} Basis states are transformed to construct mixed-state prototypes, fitting to the class quantum representations. \textbf{Stage 3: Memory Update.} Samples yielding the minimum HS distance to the prototypes are stored for future incremental tasks.}
		\label{fig:qpc_architecture}
	\end{figure*}

	\section{Mixed-state based quantum prototype classifier}
	\label{sec:Mixed-state based quantum prototype classifier}

	\subsection{Overview of the quantum prototype classifier}
	\label{subsec:Overview of the quantum prototype classifier}
	
	The core architecture of our proposed framework is a mixed-state quantum prototype classifier, as illustrated in \cref{fig:qpc_architecture}. A large class of commonly used quantum classifiers constructs class scores from a fixed set of measured basis-state probabilities or expectation values. Their measurement or readout structures may need to be expanded as the label space grows. Our design instead performs distance-based prototype matching within a shared density-matrix space.
	
	Formally, given an input instance $\mathbf{x} \in \mathcal{X}$, the quantum feature extraction backbone acts as a parameterized mapping function $F_q(\cdot; \Theta_{bb})$. It encodes and processes $\mathbf{x}$ to output an intermediate quantum mixed state, represented by a density matrix $\rho(\mathbf{x}) \in \mathbb{C}^{d \times d}$, where $d = 2^4 = 16$ for our 4-qubit subsystem. In parallel we construct an independent mixed-state quantum prototype circuit for each category $c \in \mathcal{Y}$. This circuit generates a class prototype $\rho_c$ residing in the identical $d \times d$ Hilbert space. The classification prediction $\hat{y}$ is then determined by identifying the class prototype that yields the minimum HS distance to the query instance:
	\begin{equation}
		\hat{y} = \arg\min_{c \in \mathcal{Y}} \mathcal{D}_c(\mathbf{x}) = \arg\min_{c \in \mathcal{Y}} \|\rho(\mathbf{x}) - \rho_c\|_F^2.
		\label{eq:overview_prediction}
	\end{equation}
	Then the HS distance is decomposed into a convex combination of quantum SWAP test circuits, as elaborated in \cref{subsubsec:Distance measurement via HS distance and SWAP test}.
	
	This design decouples representation learning from prototype classification. The backbone learns characteristic information, while the prototype circuit learns the characteristic distribution of specific categories within the same density matrix space. In the following subsections, we detail the construction and optimization of these core components.

	\subsection{Quantum feature extraction backbone}
	\label{subsec:Quantum feature extract backbone}
	
	The quantum feature extraction backbone serves as a hybrid classical-quantum bridge, mapping high-dimensional classical inputs into compact quantum density matrices. This backbone mainly consists of a classical pre-processing module and a subsequent quantum evolution network.
	
	Initially, a classical neural network is employed to extract high-level semantic features from the raw input data. To bridge the dimensional gap between classical feature spaces and the quantum Hilbert space, the extracted features undergo a dimension reduction and normalization process. The resulting compact vector is then embedded into the initial pure state of the quantum system using amplitude encoding, translating classical data into probability amplitudes.
	
	The embedded pure quantum state is processed by the quantum backbone, which typically employs a sequence of parameterized unitary operations interleaved with dimensionality reduction mechanisms, such as quantum pooling. The primary objective of these reduction operations is to achieve semantic concentration, extract robust features, and filter out uncorrelated noise. However, operations that discard redundant degrees of freedom, such as partial tracing, inevitably induce information loss regarding the global pure state. That means the retained subsystem is taken into a quantum mixed state.
	
	Therefore, the emergence of a mixed-state representation is a physical consequence of performing feature abstraction and denoising within a quantum architecture. The resulting density matrix $\rho(x)$ intrinsically encapsulates the structural uncertainty and hierarchical semantic information of the input, serving as a highly robust, noise-resilient foundational input for the subsequent prototype-based classification.
	
	\begin{algorithm}[!b]
		\caption{Classification via Weighted Swap Tests}
		\label{alg:classification}
		\begin{algorithmic}
			\STATE \textbf{input} $x$ \hfill // query instance
			\STATE \textbf{require} $F_q(\cdot)$ \hfill // frozen quantum feature backbone
			\STATE \textbf{require} $\mathcal{Y}_{vis}^{(t)}$ \hfill // currently visible classes
			\STATE \textbf{require} $\{U_c, \boldsymbol{w}_c\}_{c \in \mathcal{Y}_{vis}^{(t)}}$ \hfill // trained prototype parameters
			
			\STATE $\rho(x) \leftarrow F_q(x)$ \hfill // extract query density matrix
			
			\FOR{$c \in \mathcal{Y}_{vis}^{(t)}$}
			\STATE // compute weighted quantum overlap via SWAP tests:
			\STATE $\mathcal{O}_c(x) \leftarrow \sum_{i=0}^{k-1} w_{c,i} \text{Tr}\left[\rho(x) U_c|i\rangle\langle i|U_c^\dagger\right]$
			\STATE // calculate prediction logit with prototype purity offset:
			\STATE $\text{logit}_c(x) \leftarrow \mathcal{O}_c(x) - \frac{1}{2} \sum_{i=0}^{k-1} w_{c,i}^2$
			\ENDFOR
			
			\STATE $\hat{y} \leftarrow \arg\max_{c \in \mathcal{Y}_{vis}^{(t)}} \text{logit}_c(x)$ \hfill // nearest-prototype decision
			
			\STATE \textbf{output} predicted class label $\hat{y}$ and logits $L = \{\text{logit}_c(x)\}$
		\end{algorithmic}
	\end{algorithm}

	\subsection{Mixed quantum prototype preparation circuit}
	\label{subsec:mixed-quantum-prototype-prepare-circuit}
	
	For each class, the model maintains an independent parameterized prototype circuit. The purpose of this circuit is to generate a mixed quantum state in the identical density-matrix space as the backbone's quantum feature output.
	
	Following the CCPS architecture, the prototype for class $c$ is constructed as a probabilistic mixture of transformed basis states. By selecting a subset of $k$ computational basis states $|i\rangle$, the prototype density matrix $\rho_c$ is formulated as:
	\begin{equation}
		\rho_c = \sum_{i=0}^{k-1} w_{c,i} U_c |i\rangle\langle i| U_c^\dagger,
		\label{eq:prototype_rho_construction}
	\end{equation}
	where $U_c$ is the VQC for class $c$, and $w_{c,i}$ represents the corresponding weights. This convex mixture allows a single class prototype to encapsulate complex intra-class variations, thereby circumventing the representational bottleneck of forcing an entire diverse class into a rigid, singular pure state.
	
	To ensure high expressivity, the circuit $U_c$ is constructed using a hierarchical multi-layer structure. Within each layer, generic two-qubit unitary blocks are applied in an alternating entanglement topology. This configuration enables the prototype circuit to span a highly expressive transformation space, maintaining both representation capacity and parameter efficiency.
	
	A challenge in optimizing the CCPS ansatz lies in the constraints of the mixture weights $w_{c,i}$, which must satisfy the constraints of a valid probability simplex (i.e., $w_{c,i} \geq 0$ and $\sum_i w_{c,i} = 1$). The traditional approach enforces this by optimizing a subset of weights, deriving the remainder via subtraction, and applying penalty functions and truncations to prevent negative values. However, such constraints often lead to severe gradient instability and discontinuous optimization landscapes, which are detrimental to training. To address this, we introduce softmax parameterization for the mixture weights. We define a set of unconstrained trainable logits $s_{c,i} \in \mathbb{R}$ for each class, and the valid probabilities can be represented as the softmax function:
	\begin{equation}
		w_{c,i} = \frac{\exp(s_{c,i})}{\sum_{j=0}^{k-1} \exp(s_{c,j})}.
		\label{eq:softmax_weights}
	\end{equation}
	Both theory and experiments have shown that softmax parameterization has a smoother and more stable optimization process.

	\subsection{Prototype-based classification algorithm}
	\label{subsec:prototype-based-classification-algorithm}
	
	Our prototype-based classifier operates on a decoupled two-stage training paradigm. In the first stage, the quantum feature extraction backbone is optimized via an auxiliary head and subsequently frozen. In the second stage, the mixed-state prototypes are fitted without altering the backbone. During the evaluation phase, predictions are executed through a direct quantum overlap measurement.
	
	The classifier determines the label by finding the nearest mixed-state prototype $\rho_c$ among all currently visible classes $c \in \mathcal{Y}_{vis}^{(t)}$. While the objective is to minimize the HS distance $\mathcal{D}_c(x) = \|\rho(x) - \rho_c\|_F^2$. As discussed in \cref{subsubsec:Distance measurement via HS distance and SWAP test}, the measurement is reduced to a weighted sum of pure-state overlaps, since the query purity $\text{Tr}[\rho(x)^2]$ is identical across all classes, and the prototype purity $\text{Tr}[\rho_c^2] = \sum_i w_{c,i}^2$ can be pre-computed classically:
	\begin{equation}
		\mathcal{O}_c(x) \equiv \text{Tr}[\rho(x) \rho_c] = \sum_{i=0}^{k-1} w_{c,i} \text{Tr}\left[\rho(x) |\psi_{c,i}\rangle\langle\psi_{c,i}|\right],
		\label{eq:weighted_swap_test}
	\end{equation}
	where $|\psi_{c,i}\rangle = U_c|i\rangle$ is the $i$-th pure state component of the class prototype $c$. 
	
	The overlap term $\mathcal{F}_{c,i}(x) = \text{Tr}[\rho(x) |\psi_{c,i}\rangle\langle\psi_{c,i}|]$ can be efficiently and directly evaluated on quantum hardware using SWAP test. Therefore, the overall matching score for class $c$ is simply the weighted sum of these Swap Test results, using the trained classical probabilities $w_{c,i}$. 
	
	To integrate this quantum measurement with classical evaluation metrics, we formulate the classification logit as:
	\begin{equation}
		\text{logit}_c(x) = \mathcal{O}_c(x) - \frac{1}{2}\sum_{i=0}^{k-1} w_{c,i}^2,
		\label{eq:logit_formulation}
	\end{equation}
	which is mathematically equivalent to $-\frac{1}{2}\mathcal{D}_c(x)$ (ignoring the constant query purity). The final prediction selects the class with the maximum logit. This hardware-friendly, weighted SWAP test evaluation procedure is detailed in Algorithm \ref{alg:classification}.

	\section{Quantum Incremental Learning Framework}
	\label{sec:Quantum Incremental Learning Framework}
	
	Building upon the mixed-state quantum prototype classifier introduced in Section \ref{sec:Mixed-state based quantum prototype classifier}, this section details our complete incremental learning framework. The framework is designed to continuously accommodate new categories over sequential tasks while mitigating catastrophic forgetting through a density-aware replay mechanism.

	\subsection{Problem formalization}
	\label{subsec:Problem formalization}
	
	In incremental learning, the model learns from a continuous stream of data delineated by a sequence of tasks $\mathcal{T} = \{T_0, T_1, \dots, T_T\}$. At the initial task $T_0$, the model is trained on a base dataset $\mathcal{D}_0$ comprising an initial set of classes $C_0$. For each subsequent incremental task $T_t$ ($t > 0$), the model receives a new dataset $\mathcal{D}_t$ containing instances from a disjoint set of novel classes $C_t$. 
	
	At any given task $t$, the currently visible class set is the union of all classes encountered so far: $\mathcal{Y}_{vis}^{(t)} = \bigcup_{i=0}^t C_i$. Due to strict memory constraints, the model cannot access previous datasets $\{\mathcal{D}_0, \dots, \mathcal{D}_{t-1}\}$ directly. Instead, it is only permitted to maintain a limited memory buffer $\mathcal{M}$ containing a small number of exemplars from the old classes. The purpose is to incrementally update the model using only $\mathcal{D}_t \cup \mathcal{M}$, maximizing the classification accuracy over all visible classes $\mathcal{Y}_{vis}^{(t)}$ without catastrophically forgetting the old ones.

	\begin{figure}[!t]
		\centering
		\adjustbox{max width=\linewidth}{
			\begin{tikzpicture}[
				font=\footnotesize, >=stealth,
				box/.style={draw=black!80, thick, rounded corners=3pt, fill=blue!6, align=center, minimum width=2.2cm, minimum height=0.7cm},
				data/.style={draw=black!80, thick, rounded corners=3pt, fill=orange!10, align=center, minimum width=1.6cm, minimum height=0.7cm},
				model/.style={draw=black!80, thick, rounded corners=3pt, fill=green!10, align=center, minimum width=2.2cm, minimum height=0.7cm},
				arrow/.style={->, thick, draw=black!70},
				panel/.style={draw=gray!50, densely dashed, rounded corners=5pt}
				]
				
				\node[panel, minimum width=8.2cm, minimum height=6.25cm] (train_panel) at (1.4, -2.675) {};  
				\node[font=\bfseries, text=black!80, anchor=south east, inner xsep=6pt, inner ysep=4pt] at (train_panel.south east) {Incremental Training};
				
				\node[data] (dt) at (-1.3, 0) {New Data\\$\mathcal{D}_t$};
				\node[data] (et1) at (1.3, 0) {Old Memory\\$\mathcal{E}_{t-1}$};
				
				\node[box] (step1) at (0, -1.3) {\textbf{Step 1}\\Backbone Train};
				\node[model, densely dotted, fill=gray!8, text=gray!80!black] (mlp) at (3.6, -1.3) {Auxiliary MLP\\(Discarded)};
				
				\node[box] (step2) at (0, -2.6) {\textbf{Step 2}\\Prototype Train};
				\node[model, fill=red!10] (protobank) at (3.6, -2.6) {Mixed-State\\Prototypes $\rho_c$};
				
				\node[box] (step3) at (0, -3.9) {\textbf{Step 3}\\Memory Update};
				\node[data] (et) at (0, -5.2) {New Memory\\$\mathcal{E}_t$};
				
				\draw[arrow] (dt.south) -- (step1.140);
				\draw[arrow] (et1.south) -- (step1.40);
				
				\draw[arrow, densely dotted] (step1.east) -- node[above, font=\scriptsize] {$\mathcal{L}_{rep}$} (mlp.west);
				\draw[arrow] (step1.south) -- node[right, font=\scriptsize] {$\rho(\mathbf{x})$} (step2.north);
				
				\draw[<->, thick, draw=black!70] (step2.east) -- node[above, font=\scriptsize] {$\mathcal{L}_{proto}^c$} (protobank.west);
				\draw[arrow] (step2.south) -- (step3.north);
				
				\draw[arrow] (step3.south) -- (et.north);
				
				\draw[arrow, rounded corners=5pt, densely dashed, draw=blue!70!black] 
				(et.east) -- (5.0, -5.2) -- node[right, font=\scriptsize, text=blue!70!black, align=left] {Retained for\\Task $t+1$} (5.0, 0) -- (et1.east);
				
				\node[panel, minimum width=8.2cm, minimum height=3.8cm] (infer_panel) at (1.4, -7.8) {};  
				\node[font=\bfseries, text=black!80, anchor=south east, inner xsep=6pt, inner ysep=4pt] at (infer_panel.south east) {Global Inference};
				
				\node[data, fill=blue!10] (xtest) at (0, -6.5) {Query\\$\mathbf{x}_{test}$};  
				
				\node[box, fill=purple!10, minimum width=3.2cm] (step4) at (0, -7.8) {\textbf{Step 4}\\Global Distance Matching\\$\arg\min \mathcal{D}_c(\mathbf{x}_{test})$};  
				
				\node[data, fill=green!20, rounded corners=4pt] (pred) at (0, -9.1) {Prediction\\$\hat{y}$};  
				
				\draw[arrow] (xtest.south) -- (step4.north);
				\draw[arrow] (step4.south) -- (pred.north);

				\draw[arrow, rounded corners=5pt, densely dashed, draw=orange!80!black] 
				(protobank.south) -- node[right, font=\scriptsize, text=orange!80!black, align=left, pos=0.8] {Visible\\Prototypes $\mathcal{Y}_t$} (3.6, -7.8) -- (step4.east);
				
			\end{tikzpicture}
		}
		\caption[Overall flowchart of the quantum incremental learning framework.]{Overall flowchart of the quantum incremental learning framework. The process is decoupled into \textbf{incremental training phase} (Steps 1--3) and \textbf{global inference phase} (Step 4). During training, the quantum backbone is updated using new data and buffered exemplars, followed by mixed-state prototype fitting and memory updating. The auxiliary MLP head is discarded after backbone training.}
		\label{fig:incremental_flowchart}
	\end{figure}
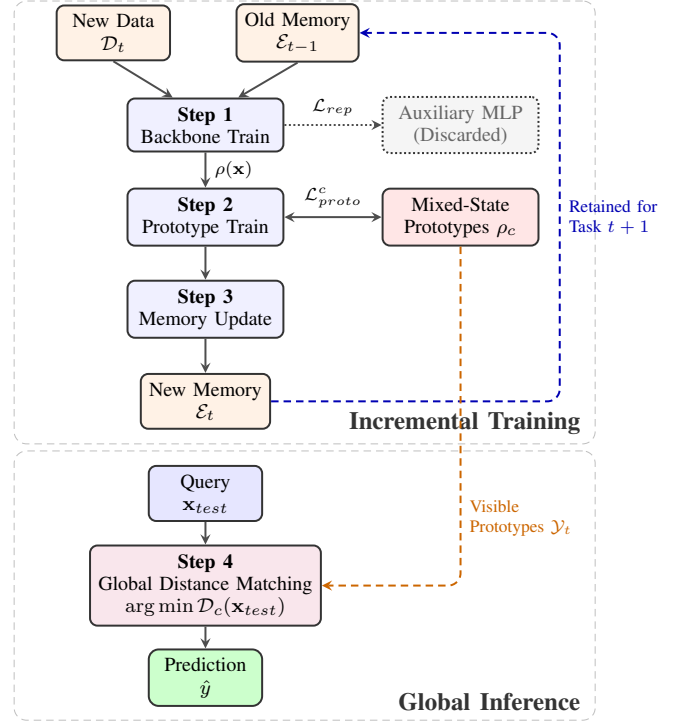

	\subsection{Details of quantum Incremental Learning Framework}
	\label{subsec:Details of quantum Incremental Learning Framework}
	
	To instantiate the theoretical model introduced in Section \ref{sec:Mixed-state based quantum prototype classifier}, we constructed the hybrid classical-quantum framework for incremental learning scenarios. This overall architecture is composed of four primary components: a classical pre-processing module, a quantum feature extraction backbone, an auxiliary MLP head for representation learning, and a set of mixed-state prototype networks.
	
	\subsubsection{Classical pre-processing module}
	The classical pre-processing module acts as the initial semantic extractor and the dimensional bridge between raw input data and the quantum computational space. Given an input instance $\mathbf{x} \in \mathcal{X}$ from the current incremental task, it is first augmented and normalized to $\mathbf{x'}$. Then the module maps the high-dimensional raw image into a dense semantic feature vector for quantum embedding.
	
	Firstly, we employ a ResNet adapted to different input sizes, denoted as $f_{res}(\cdot ; \theta_{res})$. After that, the network outputs a robust $d_{res}$-dimensional feature vector:
	\begin{equation}
		\mathbf{h} = f_{res}(\mathbf{x'}; \theta_{res}) \in \mathbb{R}^{d_{res}},
		\label{eq:resnet_feature}
	\end{equation}
	where $d_{res} = 512$ in our implementation.
	
	In order to map the dimension of $\mathbf{h}$ to that of the quantum embedding, we append a linear reducer. It compresses the 512-dim representation into $d_q$-dim, where $d_q = 2^n = 256$ as we chose the quantum circuit with width $n = 8$. The transformation is formulated as:
	\begin{equation}
		\mathbf{z} = \text{Dropout} \left( \sigma(W \mathbf{h} + \mathbf{b}) \right) \in \mathbb{R}^{d_q},
		\label{eq:linear_reduction}
	\end{equation}
	where $W \in \mathbb{R}^{256 \times 512}$ is the weight matrix, and $\mathbf{b} \in \mathbb{R}^{256}$ is the trainable bias, while $\sigma(\cdot)$ denotes the ReLU function.
	
	For quantum embedding, a valid pure quantum state must reside in a Hilbert space with a unit norm. Therefore, the $L_2$-normalized vector $\mathbf{a}$ can be represented as:
	\begin{equation}
		\mathbf{a} = \frac{\mathbf{z}}{\|\mathbf{z}\|_2}, \quad \text{s.t.} \sum_{j=0}^{2^n-1} a_j^2 = 1.
		\label{eq:l2_norm}
	\end{equation}
	Through amplitude embedding, this normalized classical vector is mapped onto the probability amplitudes of an $n$-qubit initial pure state:
	\begin{equation}
		|\psi(\mathbf{x})\rangle = \sum_{j=0}^{2^n-1} a_j |j\rangle,
		\label{eq:amplitude_embedding}
	\end{equation}
	where $|j\rangle$ represents the computational basis states spanning the $2^n$-dim Hilbert space.
	
	\subsubsection{Quantum feature extraction backbone}
	The initial pure state $|\psi(\mathbf{x})\rangle$ prepared by the classical pre-processing module serves as the direct input to the quantum feature extraction backbone. Defining the initial density matrix as $\rho_{in} = |\psi(\mathbf{x})\rangle \langle \psi(\mathbf{x})|$, the backbone adopts a quantum convolutional neural network (QCNN) architecture to distill high-level quantum semantic features\cite{cong2019quantum, du2022quantum}.

	The QCNN adopts the pyramidal structure of classical CNNs to the quantum paradigm. It alternately stacks quantum convolutional and pooling layers. In the quantum convolutional layer, the network emulates the principles of local receptive fields and weight sharing. It applies a parameterized global unitary operator $U_c(\theta_c)$, which is composed of multiple two-qubit unitary blocks acting on adjacent qubit pairs. The state evolution through the convolutional layer is governed by:
	\begin{equation}
		\rho_{conv} = U_c(\theta_c) \rho_{in} U_c^\dagger(\theta_c).
		\label{eq:quantum_conv}
	\end{equation}
	This operation entangles local qubits to capture complex spatial and feature correlations within the high-dimensional Hilbert space, mapping linearly inseparable classical data into an exponentially expansive quantum feature space. By leveraging quantum entanglement and weight sharing, the QCNN captures non-local correlations and constructs highly non-linear feature distribution. With a few dozen trainable parameters, it can avoid the parameter explosion typical of classical networks.

	Following the convolution, a quantum pooling layer is applied to compress the feature representation and improve robustness. The pooling operation first applies parameterized controlled rotations $U_p(\theta_p)$ between adjacent retained and discarded qubits, gathering information into the retained wires. Subsequently, a subset of qubits, $S$, is traced out. The retained subsystems are in a mixed state. Thus, the intermediate feature representation output by the QCNN is formulated as:
	\begin{equation}
		\rho_{pool} = \text{Tr}_S \left[ U_p(\theta_p) \rho_{conv} U_p^\dagger(\theta_p) \right].
		\label{eq:quantum_pooling}
	\end{equation}
	
	In our 8-qubit implementation, the pooling layer traces out exactly half of the qubits, reducing the system to 4 wires. Thereafter, a group of quantum convolutional layers, $U_{c2}(\theta_{c2})$, continues to be stacked on this reduced 4-qubit subsystem. The final intermediate feature representation is thus formulated as:
	\begin{equation}
		\rho(\mathbf{x}) = U_{c2}(\theta_{c2}) \rho_{pool} U_{c2}^\dagger(\theta_{c2}).
		\label{eq:quantum_conv2}
	\end{equation}
	Consequently, $\rho(\mathbf{x})$ emerges as a $16 \times 16$ complex-valued density matrix. The partial trace operation in the pooling layer naturally abstracts structural uncertainty and discards redundant noise. The quantum state after pooling is a density operator, which can characterize information uncertainty with more degrees of freedom than a pure state.

	\subsubsection{Auxiliary MLP head for representation learning} 
	To ensure that the intermediate density matrix $\rho(\mathbf{x})$ is highly discriminative over the currently visible classes, we append an auxiliary classical MLP head during the initial representation learning stage. This design is crucial when the number of visible classes $C_{vis}$ exceeds the dimension of the measured basis states (i.e., $2^4 = 16$). This scenario bottlenecks traditional explicit quantum classifiers, as the explicit method directly measures the basis-state probabilities as the logit outputs.
	
	Specifically, we perform projective measurements on the 4 retained qubits in the computational basis. The resulting 16-dim probability distribution vector $\mathbf{p}(\mathbf{x}) \in \mathbb{R}^{16}$ is extracted by calculating the diagonal elements of the output density matrix:
	\begin{equation}
		p_k(\mathbf{x}) = \text{Tr} \left[ \rho(\mathbf{x}) |k\rangle \langle k| \right], \quad k \in \{0, 1, \dots, 15\}.
		\label{eq:quantum_measurement}
	\end{equation}
	
	This 16-dimensional probability vector is then mapped to the class logits $\mathbf{\hat{y}}(\mathbf{x}) \in \mathbb{R}^{C_{vis}}$ via a two-layer MLP. The mapping is formulated as:
	\begin{equation}
		\mathbf{\hat{y}}(\mathbf{x}) = W_{out} \sigma(W_{hid} \mathbf{p}(\mathbf{x}) + \mathbf{b}_{hid}) + \mathbf{b}_{out},
		\label{eq:auxiliary_mlp}
	\end{equation}
	where $W_{hid} \in \mathbb{R}^{d_{hid} \times 16}$ and $W_{out} \in \mathbb{R}^{C_{vis} \times d_{hid}}$ are the trainable weight matrices, $\mathbf{b}_{hid}$ and $\mathbf{b}_{out}$ are the biases. In our implementation, $d_{hid}$ is set to be 32.
	
	To optimize the classical-quantum backbone, we compute the standard cross-entropy loss between the predicted logits $\mathbf{\hat{y}}(\mathbf{x})$ and the ground-truth one-hot label $\mathbf{y}_{true}$:
	\begin{equation}
		\mathcal{L}_{rep} = - \sum_{c=1}^{C_{vis}} y_{true, c} \log \left( \frac{\exp(\hat{y}_c(\mathbf{x}))}{\sum_{j} \exp(\hat{y}_j(\mathbf{x}))} \right).
		\label{eq:representation_loss}
	\end{equation}
	
	This auxiliary MLP head is utilized exclusively to provide supervised gradients for shaping a discriminative quantum density matrix space. Once the optimal feature representation is learned, this classical head is completely discarded during the subsequent mixed-state prototype fitting and decision.

	\subsubsection{Mixed-state prototype and prototype classification} 
	To align with the backbone's output, the mixed-state prototype preparation circuit is instantiated on a 4-qubit system, intrinsically operating within the identical $16 \times 16$ density-matrix space. Concretizing the CCPS ansatz, the prototype circuit adopted the hardware efficient tiling. Specifically, we constructed three layers of VQC. And each layer consists of four two-qubit $SU(4)$ blocks arranged in an alternating ring topology\cite{maccormack2022branching}. Since each $SU(4)$ block consumes exactly 15 parameters, a single class prototype circuit requires a footprint of $4 \times 3 \times 15 = 180$ trainable parameters.
	
	\begin{figure}[htbp]
		\centering
		\includegraphics[trim={0cm 0cm 0cm 0cm}, clip, width=0.98\linewidth]{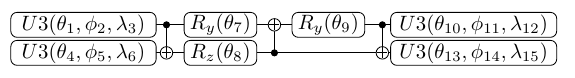}
		\caption{The internal gate decomposition of a single two-qubit $SU(4)$ block. It consumes exactly 15 trainable parameters, providing highly expressive transformations for the prototype circuit.}
		\label{fig:su4_circuit}
	\end{figure}
	
	Given the unitary operator $U_c(\theta_c)$, the circuit transforms the first $K$ ($K \le 16$) computational basis states $|i\rangle$ into a set of parameterized pure state components:
	\begin{equation}
		|\psi_{c,i}(\theta_c)\rangle = U_c(\theta_c) |i\rangle.
		\label{eq:pure_states}
	\end{equation}
	
	To form a valid mixed state, these pure states are probabilistically mixed using classical weights $w_{c,i}$. To ensure the adherence to the probability simplex constraint ($\sum w_{c,i}=1, w_{c,i} \geq 0$), we map a set of unconstrained trainable weights $s_{c,i} \in \mathbb{R}$ via the softmax function \cref{eq:softmax_weights}.

	The class-specific mixed-state prototype $\rho_c$ is thus formally assembled as the weighted sum of these orthogonal pure states:
	\begin{equation}
		\rho_c = \sum_{i=0}^{K-1} w_{c,i} |\psi_{c,i}(\theta_c)\rangle \langle \psi_{c,i}(\theta_c)|.
		\label{eq:prototype_assembly}
	\end{equation}
	
	During the prototype fitting phase, the objective is to align $\rho_c$ with the distribution of the training data. For a batch of $N$ training samples $\{\mathbf{x}_n\}_{n=1}^N$ belonging to class $c$, the quantum backbone extracts their corresponding intermediate density matrices $\{\rho_n\}_{n=1}^N$. The overlap between each sample $\rho_n$ and the prototype's pure state component $|\psi_{c,i}\rangle$ can be efficiently estimated on quantum hardware using the SWAP test:
	\begin{equation}
		\mathcal{F}_{c,i}^{(n)} = \text{Tr}\left[ \rho_n |\psi_{c,i}(\theta_c)\rangle \langle \psi_{c,i}(\theta_c)| \right].
		\label{eq:overlap_fidelity}
	\end{equation}
	
	By minimizing the mean squared Hilbert-Schmidt distance $\frac{1}{N}\sum_n \|\rho_n - \rho_c\|_F^2$ and discarding the constant target purity $\text{Tr}[\rho_n^2]$, the empirical loss function for optimizing the class prototype elegantly simplifies to:
	\begin{equation}
		\mathcal{L}_{proto}^c(\theta_c, s_c) = \sum_{i=0}^{K-1} w_{c,i}^2 - \frac{2}{N} \sum_{n=1}^N \sum_{i=0}^{K-1} w_{c,i} \mathcal{F}_{c,i}^{(n)}.
		\label{eq:prototype_loss}
	\end{equation}
	By directly optimizing this loss, the 180-parameter circuit and the $K$ weights are jointly updated to make the prototype fit the centroid of the samples. This concrete instantiation ensures robust semantic concentration with low parameter complexity, making it scalable for continuous incremental tasks.
	Let $\bar{\rho}_c=\frac{1}{N}\sum_{n=1}^{N}\rho_n$ denote the mean density matrix of class $c$. The mean squared HS distance differs from $\|\bar{\rho}_c-\rho_c\|_F^2$ only by a constant independent of $\rho_c$. Therefore, optimizing Eq. (\ref{eq:prototype_loss}) can be interpreted as applying the QMSC-based variational quantum PCA procedure to the class-level mean density matrix.
	
	\begin{algorithm}[!b]
		\caption{Quantum Incremental Learning Algorithm}
		\label{alg:qpil_algorithm}
		\begin{algorithmic}
			\STATE \textbf{input} $\mathcal{D}_1, \dots, \mathcal{D}_T$ \hfill // sequential training tasks
			\STATE \textbf{input} $M$ \hfill // total memory budget
			\STATE \textbf{require} $\Theta_{bb}$ \hfill // quantum backbone parameters
			\STATE \textbf{require} $\Theta_{head}$ \hfill // auxiliary MLP head parameters
			\STATE \textbf{require} $\mathcal{P} = \emptyset$ \hfill // mixed-state prototypes
			\STATE \textbf{require} $\mathcal{E} = \emptyset$ \hfill // exemplar memory buffer
			
			\FOR{$t = 1, \dots, T$}
			\STATE $\mathcal{Y}_t \leftarrow \text{VisibleClasses}(\mathcal{D}_1, \dots, \mathcal{D}_t)$
			\STATE $\mathcal{D}_{train} \leftarrow \mathcal{D}_t \cup \mathcal{E}$ \hfill // construct training batch
			
			\STATE // \textit{Step 1: Backbone representation learning}
			\STATE $\Theta_{bb}, \Theta_{head} \leftarrow \textsc{UpdateBackbone}(\mathcal{D}_{train}, \Theta_{bb}, \Theta_{head})$
			
			\STATE // \textit{Step 2: Mixed-state prototype fitting}
			\STATE $\mathcal{P} \leftarrow \textsc{FitMixedPrototypes}(\mathcal{D}_{train}, \mathcal{Y}_t, \Theta_{bb})$
			
			\STATE // \textit{Step 3: Replay memory management}
			\STATE $m \leftarrow \lfloor M / |\mathcal{Y}_t| \rfloor$ \hfill // number of exemplars per class
			\STATE $\mathcal{E} \leftarrow \emptyset$ \hfill // reset memory buffer
			\FOR{$c \in \mathcal{Y}_t$}
			\STATE $\mathcal{E} \leftarrow \mathcal{E} \cup \textsc{ConstructExemplars}($
			\STATE \hspace{\algorithmicindent}$\mathcal{D}_{train}, c, m, \mathcal{P}, \Theta_{bb})$
			\ENDFOR
			
			\STATE // \textit{Step 4: Prototype-based global inference}
			\FOR{$\mathbf{x}_{test} \in \mathcal{D}_{test}^{(t)}$}
			\STATE $\hat{y} \leftarrow \textsc{Classify}(\mathbf{x}_{test}, \mathcal{P}, \Theta_{bb})$
			\ENDFOR
			
			\ENDFOR
		\end{algorithmic}
	\end{algorithm}

	\subsection{Quantum incremental learning algorithm}
	\label{subsec:Quantum incremental learning algorithm}
	The quantum incremental learning algorithm is a multi-stage process. Assuming a continuous data stream divided into sequential tasks $t \in \{0, 1, \dots, T\}$, let $\mathcal{Y}_t$ denote the set of all visible classes up to task $t$. At task $t$, the model receives a new dataset $\mathcal{D}_t$ and has access to a limited replay memory $\mathcal{E}_{t-1}$ containing exemplars from old classes. The incremental learning procedure follows a four-step algorithmic pipeline, which is summarized in \cref{alg:qpil_algorithm}.
	
	\subsubsection{Step 1: Backbone representation learning} 
	The training batch is constructed as $\mathcal{D}_{train}^{(t)} = \mathcal{D}_t \cup \mathcal{E}_{t-1}$. The classical pre-processing module and the quantum feature extraction backbone are jointly updated. During this phase, the intermediate density matrix $\rho(\mathbf{x})$ is mapped to class logits via the auxiliary MLP head. The entire backbone is optimized by minimizing loss $\mathcal{L}_{rep}$ (Eq. \ref{eq:representation_loss}) over all visible classes in $\mathcal{Y}_t$. 
	
	\subsubsection{Step 2: Mixed-state prototype fitting} 
	Once the shared representation is learned, the auxiliary MLP head is detached. For each class $c \in \mathcal{Y}_t$, the parameters $(\theta_c, s_c)$ of the class-specific prototype circuit are then independently optimized by minimizing loss $\mathcal{L}_{proto}^c$ (Eq. \ref{eq:prototype_loss}), guided by the hardware-efficient SWAP test.
	
	\subsubsection{Step 3: Replay memory management} 
	After prototype fitting, the exemplar set $\mathcal{E}_{t}$ is updated. For each class, evaluate the squared HS distance between the sample density matrix $\rho(\mathbf{x})$ and the fitted prototype $\rho_c$ via SWAP test. Samples yielding the minimum distance are stored in $\mathcal{E}_{t}$. The memory budget is distributed equally across classes.
	
	\subsubsection{Step 4: Prototype-based global inference} 
	Any query instance $\mathbf{x}_{\text{test}}$ is passed through the frozen backbone to extract its density matrix $\rho(\mathbf{x}_{\text{test}})$ in the form of quantum states. The prediction is determined by identifying the nearest mixed-state prototype across all visible classes $\mathcal{Y}_t$:
	\begin{equation}
		\hat{y} = \arg\min_{c \in \mathcal{Y}_t} \mathcal{D}_c(\mathbf{x}_{test}).
		\label{eq:final_prediction}
	\end{equation}
	In practical implementation, the negative distance $-\mathcal{D}_c(\mathbf{x}_{test})$ directly serves as the output logit.
	
	\section{Experiment}
	\label{sec:Experiment}
	
	\subsection{Experimental Setup}
	\label{subsec:experimental_setup}
	
	\subsubsection{Datasets}To evaluate the performance of the proposed framework, we conduct experiments on CIFAR-100 and TinyImageNet.
	
	CIFAR-100: It consists of 60,000 RGB color images with a spatial resolution of $32 \times 32$ pixels. The dataset contains 100 classes, representing a diverse set of everyday objects. For each class, it is divided into 500 images for training and 100 for testing. To evaluate the sensitivity to class selection, we construct three fixed and mutually disjoint 32-class subsets, denoted as Split A, Split B, and Split C, which cover 96 classes in total.
	
	TinyImageNet: This dataset serves as a scaled-down modification of the full ImageNet dataset, frequently utilized to bridge the gap between low-resolution datasets and large-scale images. It comprises 200 classes. Each class contains 500 training images, 50 validation images, and 50 testing images, all possessing a spatial resolution of $64 \times 64$ pixels.
	
	For each CIFAR-100 split and TinyImageNet, we selected 32 categories and divided them into 16 initial classes and 4 groups of 4 categories for incremental tasks. Experiments resulted in five tasks from the 16 to 32 classes. These splits were then fixed in the experiment. The detailed dataset split protocol is provided in the source code.

	\subsubsection{Experimental configuration} This study utilizes the PyTorch and PennyLane frameworks for hybrid model construction and quantum circuit simulation. All experiments are executed on an x86 platform equipped with a single NVIDIA GeForce RTX 3090 (24GB) GPU. In the statistical significance experiments, we randomly select four seeds and report the results as mean $\pm$ standard deviation. For all other experiments, the random seed is fixed to 1993. All experiments used a validation split of 10\%. We used ResNet-18 preprocessing adapted to the image resolution, followed by a 256-dimensional MLP projection and an 8-qubit quantum backbone. To avoid dataset-specific hyperparameter selection, we fix the prototype rank to $K=7$ for every dataset and class split. This value is specified uniformly before the comparative evaluation and is not selected according to test-set performance on individual datasets. The initial task was trained for 240 epochs with a batch size of 128 and a learning rate of 0.01. Incremental tasks used prototype-nearest exemplar replay with a fixed memory budget of 640 samples, allocated in a per-class manner (approximately 20 samples per class after all 32 classes are observed). The prototype branch was trained for 60 epochs with a learning rate of 0.04 and a density batch size of 16, while the incremental finetuning stage was trained for 36 epochs with a learning rate of 0.01. We used SGD with momentum 0.9 and weight decay 1e-4 for backbone/classifier training, and Adam for prototype optimization. A warm-up cosine scheduler was used for the initial task, and a cosine scheduler was used for incremental finetuning. The source code is available at: \url{https://anonymous.4open.science/r/QPIL-3E43}.
	
	\subsubsection{Quantum circuit configuration} Due to the constraints of current quantum resources, the quantum classification backbone is implemented with an 8-qubit circuit, while the mixed-state prototype circuit for each class is instantiated on a 4-qubit system. We use the HS distance between density matrices, rather than constructing an entire 13-qubit circuit, when the SWAP test will also introduce an auxiliary qubit. By disassembling the modules, a significant amount of storage was saved. The 8-qubit quantum classifier exposes an intermediate 4-qubit density matrix on the retained wires after the partial QCNN transformation, and each CCPS prototype is also represented as a $16 \times 16$ density matrix in the same 4-qubit Hilbert space.

	\subsubsection{Evaluation Metrics}
	\label{sec:metrics}
	We evaluate our method and the baselines using two primary metrics: average accuracy (Avg Acc) and last accuracy (Last Acc). 
	
	Let $T$ be the total number of incremental tasks. We denote $A_t$ as the top-1 classification accuracy evaluated on the test set of all seen classes from task $1$ to task $t$, after the model has finished training on task $t$. 
	
	Last Acc: This metric measures the final performance of the model after the entire sequence of incremental learning is completed. It is defined as the accuracy on all learned classes after the last task $T$:
	\begin{equation}
		\mathcal{A}_{last} = A_T
	\end{equation}
	
	Avg Acc: This metric reflects the historical performance and the robustness of the model against catastrophic forgetting throughout the entire continual learning process. It is defined as the average of the top-1 incremental accuracies over all $T$ tasks:
	\begin{equation}
		\mathcal{A}_{avg} = \frac{1}{T} \sum_{t=1}^{T} A_t
	\end{equation}

	\begin{table*}[!t]
		\centering
		\caption{Comparison with classical and quantum classifiers.}
		\label{tab:quantum_prototype_effectiveness}
		\scriptsize
		\setlength{\tabcolsep}{1.5pt}
		\begin{tabular*}{\textwidth}{@{\extracolsep{\fill}}llccccc}
			\toprule
			Type & Classifier & Split A & Split B & Split C & TinyImageNet & Avg \\
			\midrule
			Classical & Linear Softmax & 0.8281 $\pm$ 0.0062 & 0.7895 $\pm$ 0.0046 & \textbf{0.8245 $\pm$ 0.0059} & 0.6844 $\pm$ 0.0100 & 0.7816 $\pm$ 0.0062 \\
			Classical & Cosine Classifier & 0.8259 $\pm$ 0.0091 & \textbf{0.7929 $\pm$ 0.0062} & 0.8223 $\pm$ 0.0040 & 0.6783 $\pm$ 0.0043 & 0.7798 $\pm$ 0.0037 \\
			Classical & Nearest Centroid & 0.8280 $\pm$ 0.0042 & 0.7884 $\pm$ 0.0045 & 0.8240 $\pm$ 0.0049 & 0.6845 $\pm$ 0.0094 & 0.7813 $\pm$ 0.0052 \\
			Classical & RBF-SVM & 0.8257 $\pm$ 0.0045 & 0.7909 $\pm$ 0.0073 & 0.8233 $\pm$ 0.0050 & 0.6881 $\pm$ 0.0109 & 0.7820 $\pm$ 0.0048 \\
			\midrule
			Quantum & HEA-VQC \cite{schuld2020circuit} & 0.6601 $\pm$ 0.0038 & 0.6066 $\pm$ 0.0105 & 0.6484 $\pm$ 0.0066 & 0.5517 $\pm$ 0.0129 & 0.6167 $\pm$ 0.0064 \\
			Quantum & QCNN \cite{cong2019quantum} & 0.6741 $\pm$ 0.0293 & 0.6259 $\pm$ 0.0161 & 0.6781 $\pm$ 0.0479 & 0.5709 $\pm$ 0.0458 & 0.6372 $\pm$ 0.0342 \\
			Quantum & TTN-QNN \cite{grant2018hierarchical} & 0.1089 $\pm$ 0.0317 & 0.1341 $\pm$ 0.0218 & 0.1346 $\pm$ 0.0244 & 0.1166 $\pm$ 0.0322 & 0.1236 $\pm$ 0.0119 \\
			Quantum & Fidelity Classifier \cite{stein2022quclassi} & 0.7311 $\pm$ 0.0436 & 0.7584 $\pm$ 0.0247 & 0.7965 $\pm$ 0.0295 & 0.6586 $\pm$ 0.0469 & 0.7362 $\pm$ 0.0194 \\
			Quantum & Mixed-state Prototype Classifier (Ours) & \textbf{0.8323 $\pm$ 0.0082} & 0.7766 $\pm$ 0.0165 & 0.8222 $\pm$ 0.0117 & \textbf{0.6997 $\pm$ 0.0106} & \textbf{0.7827 $\pm$ 0.0033} \\
			\bottomrule
		\end{tabular*}
		\\[2pt]
		\raggedright\scriptsize{Avg denotes the average accuracy over the four datasets.}
	\end{table*}

	\begin{table*}[!t]
		\centering
		\caption{Class-incremental comparison with replay-based methods.}
		\label{tab:class_incremental_results}
		\scriptsize
		\setlength{\tabcolsep}{1.5pt}
		\begin{tabular*}{\textwidth}{@{\extracolsep{\fill}}lllcccccccc}
			\toprule
			\multirow{2}{*}{Model}
			& \multirow{2}{*}{Memory}
			& \multirow{2}{*}{\makecell{Parameter\\magnitude}}
			& \multicolumn{2}{c}{Split A}
			& \multicolumn{2}{c}{Split B}
			& \multicolumn{2}{c}{Split C}
			& \multicolumn{2}{c}{TinyImageNet} \\
			\cmidrule(lr){4-5} \cmidrule(lr){6-7} \cmidrule(lr){8-9} \cmidrule(lr){10-11}
			& & & Last & Avg & Last & Avg & Last & Avg & Last & Avg \\
			\midrule
			FOSTER\cite{wang2022foster} & 640 & $\mathcal{O}(nd)$ & 0.6847 & 0.7796 & 0.5850 & 0.7006 & 0.6722 & 0.7711 & 0.3588 & 0.4756 \\
			iCaRL\cite{rebuffi2017icarl} & 2000 & $\mathcal{O}(nd)$ & 0.7170 & 0.7754 & 0.6690 & 0.7468 & 0.7280 & 0.7960 & 0.5780 & 0.6488 \\
			PODNet\cite{douillard2020podnet} & 2000 & $\mathcal{O}(nmd)$ & 0.6940 & 0.7746 & 0.6480 & 0.7362 & 0.6840 & 0.7760 & 0.5440 & 0.6194 \\
			\textbf{Ours} ($K=7$) & 640 & $\mathcal{O}(n\log_{2}(\max(n,d)))$ & 0.5728 & 0.7178 & 0.4947 & 0.6465 & 0.5481 & 0.7097 & 0.3838 & 0.5371 \\
			BiC\cite{wu2019large} & 640 & $\mathcal{O}(nd)$ & 0.5308 & 0.6719 & 0.1997 & 0.4802 & 0.2578 & 0.5848 & 0.0511 & 0.2524 \\
			SSRE\cite{zhu2022self} & 2000 & $\mathcal{O}(nd)$ & 0.4947 & 0.5970 & 0.4522 & 0.5776 & 0.5025 & 0.6276 & 0.4319 & 0.5259 \\
			DER\cite{buzzega2020dark} & 640 & $\mathcal{O}(nd)$ & 0.3941 & 0.5785 & 0.5374 & 0.6466 & 0.5603 & 0.6996 & 0.0313 & 0.1551 \\
			LUCIR\cite{hou2019learning} & 2000 & $\mathcal{O}(nd)$ & 0.1690 & 0.2206 & 0.2090 & 0.2486 & 0.1800 & 0.2394 & 0.1800 & 0.2178 \\
			\bottomrule
		\end{tabular*}\\[2pt]
		\raggedright\scriptsize{Memory denotes the maximum number of retained real training samples. $n$: number of classes; $d$: feature dimension; $m$: number of proxies per class.}
	\end{table*}

	\subsection{Effectiveness of the Quantum Prototype Classifier}
	\label{subsec:effectiveness_quantum_prototype_classifier}

	We evaluate the effectiveness of the mixed-state quantum prototype classifier proposed in \cref{sec:Mixed-state based quantum prototype classifier}. The purpose of this experiment is to compare the proposed prototype-based decision mechanism with simple classical classifiers and representative quantum classifiers.

	We compare our method with four simple classical classifiers and four quantum classifiers. Linear Softmax and Cosine Classifier use parametric classification heads. Nearest Centroid and RBF-SVM are fitted on normalized features extracted by independently trained backbones. The quantum baselines retain the ResNet-18 preprocessing and use 8-qubit amplitude encoding. HEA-VQC applies RY-RZ layers and ring CNOT gates, followed by direct five-qubit probability measurement for 32-class prediction \cite{schuld2020circuit}. QCNN uses SU(4) convolution and quantum pooling, followed by the same direct probability measurement \cite{cong2019quantum}. TTN-QNN uses a hierarchical 8-qubit tree circuit and a 16-to-32 classical readout layer \cite{grant2018hierarchical}. Fidelity Classifier represents each class using a trainable pure-state prototype and performs classification based on state fidelity \cite{stein2022quclassi}. Our method instead represents each class using a mixed-state prototype and performs classification based on the HS distance. All methods are evaluated under the same non-incremental setting on the same 32 selected classes. The results are reported as mean $\pm$ standard deviation over four runs.

	As shown in \cref{tab:quantum_prototype_effectiveness}, our method achieves mean accuracies of 0.8323, 0.7766, 0.8222, and 0.6997 on the four datasets. It obtains the highest mean accuracy among the compared quantum classifiers on all four datasets. It also achieves the highest accuracy among all compared classifiers on Split A and TinyImageNet. On Split B and Split C, it is 1.63 and 0.23 percentage points lower than the strongest classical classifier, respectively. Its average accuracy over the four datasets is 0.7827, which is the highest among the compared methods. Compared with the direct-measurement QCNN, our method improves the mean accuracy by 15.83, 15.07, 14.41, and 12.88 percentage points, respectively. These results demonstrate that directly relying on the measured output of the quantum backbone is insufficient for robust multi-class classification because the measured probability vector is constrained by the retained quantum basis states. It assigns the same fixed Hilbert space size to each category and therefore cannot effectively handle uneven feature distributions. In contrast, our approach is to adaptively divide the feature space into different ranges for different classes.

	\subsection{Effectiveness on Incremental Learning}
	\label{subsec:class_incremental_effectiveness}

	This section evaluates the effectiveness of the proposed method under the incremental learning setting. The purpose is not to claim state-of-the-art performance over all classical methods, but to verify whether the proposed quantum method has an effective incremental learning capability. Since the model operates with a limited-width quantum circuit and a fixed memory budget, the key question is whether it can alleviate catastrophic forgetting.

	We follow the fixed incremental protocol introduced in \cref{subsec:experimental_setup}. The model starts from 16 initial classes and then learns four incremental tasks, each containing 4 new classes, resulting in five evaluation stages from 16 to 32 classes. For comparison, we organize the representative baselines into four categories. Replay-based methods include FOSTER, iCaRL, and PODNet \cite{wang2022foster, rebuffi2017icarl, douillard2020podnet}; BiC, SSRE, and DER \cite{wu2019large, zhu2022self, buzzega2020dark}; and LUCIR \cite{hou2019learning}. Prototype-based methods include FeTrIL, PASS, and NCM \cite{petit2023fetril, zhu2021prototype, mensink2012metric}, as well as Eucl-NCM and TOPIC+ \cite{goswami2023fecam, tao2020few}. Statistics-based methods include FeCAM, SDC, and DeeSIL \cite{goswami2023fecam, yu2020semantic, belouadah2018deesil}. Regularization-based methods include ABD, MUC-LwF, and LwF-MC \cite{smith2021always, liu2020more, li2017learning}, as well as EWC \cite{kirkpatrick2017overcoming}. Finetune is included as a conventional sequential baseline. Our method combines mixed-state prototype classification with exemplar replay and is therefore included in both tables as a common reference. All baseline results were reproduced using their official open-source implementations. The same class subsets and class order were used for all methods. Each baseline used either the same ResNet backbone as ours or the backbone provided by its official implementation. All methods are evaluated by the final accuracy after the last task, denoted as Last Acc, and the average accuracy over all incremental stages, denoted as Avg Acc.

	\begin{table*}[!t]
		\centering
		\caption{Class-incremental comparison with prototype, statistics, and regularization methods.}
		\label{tab:class_incremental_prototype_results}
		\scriptsize
		\setlength{\tabcolsep}{1.5pt}
		\begin{tabular*}{\textwidth}{@{\extracolsep{\fill}}lllcccccccc}
			\toprule
			\multirow{2}{*}{Model}
			& \multirow{2}{*}{Type}
			& \multirow{2}{*}{\makecell{Parameter\\magnitude}}
			& \multicolumn{2}{c}{Split A}
			& \multicolumn{2}{c}{Split B}
			& \multicolumn{2}{c}{Split C}
			& \multicolumn{2}{c}{TinyImageNet} \\
			\cmidrule(lr){4-5} \cmidrule(lr){6-7} \cmidrule(lr){8-9} \cmidrule(lr){10-11}
			& & & Last & Avg & Last & Avg & Last & Avg & Last & Avg \\
			\midrule
			FeCAM\cite{goswami2023fecam} & Statistics & $\mathcal{O}(nd^{2})$ & 0.6869 & 0.7826 & 0.6181 & 0.7330 & 0.6728 & 0.7785 & 0.5919 & 0.6653 \\
			FeTrIL\cite{petit2023fetril} & Prototype & $\mathcal{O}(nd)$ & 0.6453 & 0.7417 & 0.5462 & 0.6613 & 0.6319 & 0.7474 & 0.5544 & 0.6291 \\
			PASS\cite{zhu2021prototype} & Prototype & $\mathcal{O}(4nd)$ & 0.6347 & 0.7236 & 0.5822 & 0.7073 & 0.5984 & 0.7142 & 0.4538 & 0.5579 \\
			\textbf{Ours} ($K=7$) & Prototype & $\mathcal{O}(n\log_{2}(\max(n,d)))$ & 0.5728 & 0.7178 & 0.4947 & 0.6465 & 0.5481 & 0.7097 & 0.3838 & 0.5371 \\
			ABD\cite{smith2021always} & Regularization & $\mathcal{O}(nd)$ & 0.5445 & 0.6487 & 0.5218 & 0.6431 & 0.5544 & 0.6976 & 0.4815 & 0.6076 \\
			NCM\cite{mensink2012metric} & Prototype & $\mathcal{O}(nd)$ & 0.4863 & 0.6598 & 0.4438 & 0.6115 & 0.4700 & 0.6519 & 0.4431 & 0.5962 \\
			MUC-LwF\cite{liu2020more} & Regularization & $\mathcal{O}(nd)$ & 0.4925 & 0.6299 & 0.4178 & 0.5889 & 0.4863 & 0.6643 & 0.3069 & 0.4981 \\
			Eucl-NCM\cite{goswami2023fecam} & Prototype & $\mathcal{O}(nd)$ & 0.4612 & 0.6291 & 0.5475 & 0.6918 & 0.5600 & 0.7127 & 0.5075 & 0.6306 \\
			Finetune & Baseline & $\mathcal{O}(nd)$ & 0.3459 & 0.5212 & 0.3378 & 0.5202 & 0.3869 & 0.5838 & 0.2856 & 0.4764 \\
			DeeSIL\cite{belouadah2018deesil} & Statistics & $\mathcal{O}(nd)$ & 0.2553 & 0.4904 & 0.2078 & 0.4555 & 0.2534 & 0.5191 & 0.1988 & 0.3597 \\
			LwF-MC\cite{li2017learning} & Regularization & $\mathcal{O}(nd)$ & 0.4290 & 0.5319 & 0.3914 & 0.5280 & 0.3513 & 0.5451 & 0.1412 & 0.3366 \\
			TOPIC+\cite{tao2020few} & Prototype & $\mathcal{O}(nd)$ & 0.1597 & 0.3741 & 0.1578 & 0.3763 & 0.1159 & 0.3856 & 0.2056 & 0.4151 \\
			SDC\cite{yu2020semantic} & Statistics & $\mathcal{O}(nd)$ & 0.1403 & 0.3566 & 0.1225 & 0.3286 & 0.0997 & 0.3546 & 0.1213 & 0.2972 \\
			EWC\cite{kirkpatrick2017overcoming} & Regularization & $\mathcal{O}(nd)$ & 0.0259 & 0.1987 & 0.1805 & 0.3931 & 0.1525 & 0.3972 & 0.0273 & 0.1592 \\
			\bottomrule
		\end{tabular*}\\[2pt]
		\raggedright\footnotesize{$n$: number of classes; $d$: feature dimension; $m$: number of proxies per class.}
	\end{table*}

	\begin{figure*}[!t]
		\centering
		\includegraphics[width=\textwidth]{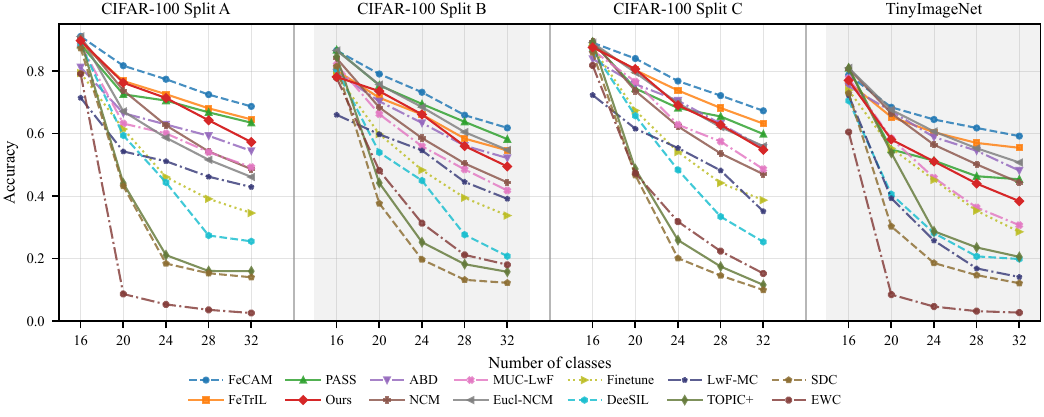}
		\caption{Accuracy at each incremental stage for the prototype, statistics, regularization, and finetuning methods in Table III. Solid, dashed, dash-dotted, and dotted lines denote prototype, statistics, regularization, and baseline methods, respectively.}
		\label{fig:table3_incremental_comparison}
	\end{figure*}

	As shown in \cref{tab:class_incremental_results,tab:class_incremental_prototype_results}, with the same fixed rank $K=7$ on all datasets, the proposed method achieves 0.5728 Last Acc and 0.7178 Avg Acc on CIFAR-100 Split A. Although it does not outperform the strongest classical baselines, its average accuracy is higher than that of BiC, Eucl-NCM, NCM, MUC-LwF, SSRE, DeeSIL, TOPIC+, SDC, and LUCIR. The corresponding Avg Acc values are 0.6465 and 0.7097 on Split B and Split C, respectively, showing that the method remains sensitive to class selection even under a uniform rank. On TinyImageNet, the same $K=7$ configuration obtains 0.3838 Last Acc and 0.5371 Avg Acc. The stage-wise curves in \cref{fig:table3_incremental_comparison} further show that the proposed method exhibits a gradual accuracy decline comparable to several prototype and statistics baselines, rather than an abrupt collapse. These results demonstrate meaningful class-incremental learning capability without selecting a separate rank for each benchmark.

	More importantly, this result is achieved under a compact quantum setting. The quantum backbone only uses 8 qubits, and the prototypes use 4 qubits. The trainable quantum backbone contains only a small number of quantum parameters, while each class prototype uses 180 circuit parameters and $K$ mixture weights. Despite this limited quantum width and fixed memory budget, the model shows a gradual decline in accuracy over the five incremental stages. These results suggest that mixed-state quantum prototypes and prototype-guided replay can mitigate forgetting to a certain extent and provide a feasible direction for quantum incremental learning under NISQ-era resource constraints.

	\subsection{Ablation Studies}
	\label{subsec:ablation_studies}

	We conduct ablation studies on two core designs: the mixed-state prototype representation and the exemplar replay mechanism. The former is used to improve the representation capacity of class prototypes by modeling each class as a density matrix rather than a single pure state. The latter is used to preserve old-class information during incremental updates by replaying a small number of representative exemplars.

	For the mixed-state prototype ablation, we replace the proposed CCPS mixed-state prototype with a pure-state prototype, denoted as w/o Mixed State. This variant keeps the same incremental learning protocol but represents each class using a single pure state. For the backbone ablation, we freeze the shared backbone and only fit new-class prototypes, denoted as Fixed Backbone. For the replay ablation, we remove the exemplar replay mechanism, denoted as w/o Replay. This variant updates the model using only new-task data.

	\begin{table*}[!t]
		\centering
		\caption{Ablation study of mixed-state prototypes and exemplar replay.}
		\label{tab:ablation_study}
		\footnotesize
		\setlength{\tabcolsep}{3.5pt}
		\begin{tabular*}{\textwidth}{@{\extracolsep{\fill}}lcccccccc}
			\toprule
			\multirow{2}{*}{Method}
			& \multicolumn{2}{c}{Split A}
			& \multicolumn{2}{c}{Split B}
			& \multicolumn{2}{c}{Split C}
			& \multicolumn{2}{c}{TinyImageNet} \\
			\cmidrule(lr){2-3} \cmidrule(lr){4-5} \cmidrule(lr){6-7} \cmidrule(lr){8-9}
			& Last & Avg & Last & Avg & Last & Avg & Last & Avg \\
			\midrule
			w/o Mixed State & 0.5450 & 0.6668 & 0.4156 & 0.5634 & 0.4772 & 0.6288 & 0.3656 & 0.4994 \\
			Fixed Backbone & 0.4788 & 0.6586 & 0.3997 & 0.5624 & 0.4559 & 0.6355 & 0.3769 & 0.5242 \\
			w/o Replay & 0.5122 & 0.6735 & 0.4425 & 0.5931 & 0.4772 & 0.6509 & 0.3850 & 0.5202 \\
			Ours (full, $K=7$) & \textbf{0.5728} & \textbf{0.7178} & \textbf{0.4947} & \textbf{0.6465} & \textbf{0.5481} & \textbf{0.7097} & \textbf{0.3838} & \textbf{0.5371} \\
			\bottomrule
		\end{tabular*}
	\end{table*}
	
\begin{figure}[H]
		\centering
		\includegraphics[trim={0cm 0.01cm 0cm 0cm},clip,width=\linewidth]{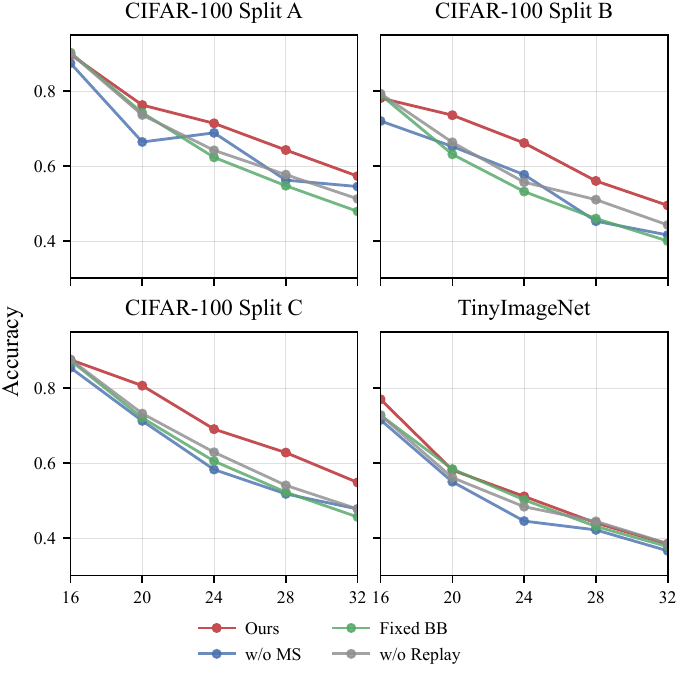}
		\caption{Accuracy at each incremental stage for different ablation variants on CIFAR-100 Split A, Split B, Split C, and TinyImageNet.}
		\label{fig:ablation_study}
	\end{figure}

	As shown in \cref{tab:ablation_study,fig:ablation_study}, on CIFAR-100 Split A, replacing the fixed-rank ($K=7$) mixed-state prototype with a pure-state prototype leads to a clear performance drop. The Last Acc decreases from 0.5728 to 0.5450, and the Avg Acc drops from 0.7178 to 0.6668. This demonstrates that a single pure state is not sufficient to represent complex intra-class variations in the incremental setting. In contrast, the proposed mixed-state prototype uses a convex combination of multiple pure-state components, which provides a more flexible representation for each class. The full model also retains the highest Avg Acc among the compared variants on Split B, Split C, and TinyImageNet.
	
	The Fixed Backbone variant obtains only 0.4788 Last Acc and 0.6586 Avg Acc on Split A. The w/o Replay variant improves the result to 0.5122 Last Acc and 0.6735 Avg Acc, but it still performs worse than the fixed-$K=7$ full model. This shows that directly updating the model with only new-task data is insufficient for maintaining a stable prototype space. By replaying exemplars nearest to the learned quantum prototypes, the full method provides old-class constraints during incremental training and helps different prototype positions remain appropriately matched.

	\subsection{Effect of Prototype Rank}
	\label{subsec:effect_quantum_pca}
	
	This section further investigates the PCA-like property of the proposed mixed-state prototype. As discussed in the previous sections, the CCPS prototype represents each class as a convex combination of multiple transformed basis states. The number of retained basis states $K$ controls the effective rank of the prototype density matrix. Therefore, $K$ plays a role similar to the number of principal components in PCA: a small $K$ may discard useful class information, while an excessively large $K$ may introduce redundant or noisy components. This experiment evaluates whether an appropriate rank selection can improve the robustness of the mixed-state prototype.
	
	We conduct a rank sweep by varying $K$ in the CCPS prototype while keeping the other incremental learning settings unchanged. Specifically, we evaluate $K$ from 1 to 16 under the same 16-to-32 class-incremental protocol. When $K=16$, the prototype uses all basis states in the 4-qubit Hilbert space, corresponding to a full-rank representation. Smaller values of $K$ force the prototype to retain only a subset of basis components, which can be regarded as a low-rank approximation of the class density distribution.
	
	\begin{table*}[htbp]
		\centering
		\caption{Sensitivity of class-incremental learning to Prototype Rank $K$.}
		\label{tab:quantum_pca_rank_sweep}
		\footnotesize
		\setlength{\tabcolsep}{3.5pt}
		\begin{tabular*}{\textwidth}{@{\extracolsep{\fill}}lcccccccc}
			\toprule
			\multirow{2}{*}{$K$}
			& \multicolumn{2}{c}{Split A}
			& \multicolumn{2}{c}{Split B}
			& \multicolumn{2}{c}{Split C}
			& \multicolumn{2}{c}{TinyImageNet} \\
			\cmidrule(lr){2-3} \cmidrule(lr){4-5} \cmidrule(lr){6-7} \cmidrule(lr){8-9}
			& Last & Avg & Last & Avg & Last & Avg & Last & Avg \\
			\midrule
			1  & 0.5184 & 0.6913 & 0.4103 & 0.5567 & 0.4834 & 0.6369 & 0.3656 & 0.4994 \\
			2  & 0.5372 & 0.6758 & 0.4706 & 0.5848 & 0.4400 & 0.6197 & 0.3600 & 0.5043 \\
			3  & 0.5459 & 0.6995 & 0.4603 & 0.6104 & 0.5072 & 0.6590 & 0.3844 & 0.5122 \\
			4  & 0.5516 & 0.7066 & 0.4900 & 0.6311 & 0.5122 & 0.6880 & \textbf{0.3931} & 0.5212 \\
			5  & 0.5972 & 0.7193 & 0.4872 & 0.6254 & 0.5063 & 0.6797 & 0.3744 & 0.5124 \\
			6  & 0.6009 & 0.7273 & 0.4881 & 0.6397 & 0.5291 & 0.6989 & 0.3788 & 0.5148 \\
			\textbf{7}  & 0.5728 & 0.7178 & 0.4947 & \textbf{0.6465} & 0.5481 & 0.7097 & 0.3838 & \textbf{0.5371} \\
			8  & 0.6000 & 0.7260 & 0.4831 & 0.6325 & 0.5547 & 0.7142 & 0.3588 & 0.5049 \\
			9  & 0.5756 & 0.7215 & \textbf{0.4988} & 0.6436 & 0.5513 & 0.7122 & 0.3819 & 0.5160 \\
			10 & 0.6016 & 0.7232 & 0.4919 & 0.6255 & 0.5253 & 0.6798 & 0.3606 & 0.5023 \\
			11 & 0.6050 & 0.7271 & 0.4913 & 0.6285 & 0.5603 & 0.7118 & 0.3744 & 0.5282 \\
			12 & \textbf{0.6144} & \textbf{0.7286} & 0.4813 & 0.6205 & 0.5469 & 0.6950 & 0.3750 & 0.5191 \\
			13 & 0.5647 & 0.7132 & 0.4781 & 0.6337 & 0.5391 & 0.6781 & 0.3594 & 0.5082 \\
			14 & 0.5919 & 0.7148 & 0.4938 & 0.6331 & 0.5563 & 0.6953 & 0.3388 & 0.5102 \\
			15 & 0.5853 & 0.7126 & 0.4966 & 0.6370 & \textbf{0.5741} & \textbf{0.7159} & 0.3744 & 0.5154 \\
			16 & 0.5731 & 0.7069 & 0.4897 & 0.6321 & 0.5656 & 0.7080 & 0.3650 & 0.5173 \\
			\bottomrule
		\end{tabular*}
	\end{table*}
	
	As shown in \cref{tab:quantum_pca_rank_sweep}, the performance does not monotonically improve as $K$ increases. Although $K=16$ provides the largest representational capacity, it does not consistently obtain the best Avg Acc. This suggests that using all basis components may preserve redundant or noisy variations in the class density matrix, which weakens the robustness of the prototype. In contrast, moderate-rank settings generally achieve better performance. The retrospective sweep reaches its best Avg Acc at $K=12$, $K=7$, and $K=15$ on Split A, Split B, and Split C, respectively, while TinyImageNet reaches its best Avg Acc at $K=7$. Importantly, these sweep maxima are not used to choose a separate setting for each dataset: all main results in the preceding sections use the same prespecified value, $K=7$. The sweep is reported only as a sensitivity analysis. This protocol avoids selecting the rank according to test performance and provides a consistent basis for cross-dataset comparison.
	
\begin{figure}[H]
		\centering
		\includegraphics[width=\linewidth]{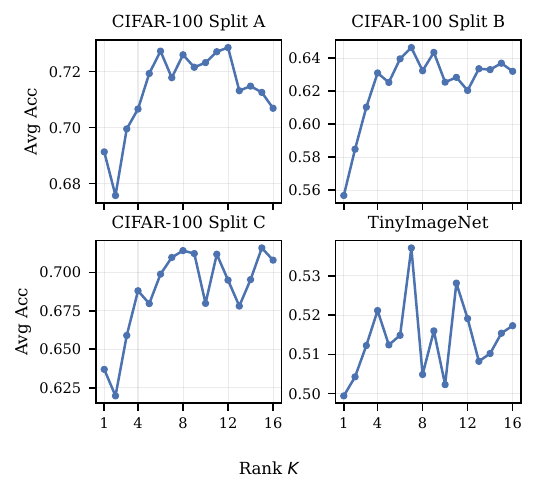}
		\caption{Effect of Prototype Rank $K$ on Avg Acc for CIFAR-100 Split A, Split B, Split C, and TinyImageNet.}
		\label{fig:quantum_pca_stability}
	\end{figure}

	{\looseness=-1
	\cref{fig:quantum_pca_stability} further illustrates the influence of $K$ from the perspective of stability. When $K$ is too small, the prototype only keeps a few dominant components, which may remove noise but also risks discarding useful intra-class variations. When $K$ is too large, especially at the full-rank setting $K=16$, the prototype absorbs more low-contribution components, increasing the fluctuation across incremental tasks and reducing the denoising effect. The middle range of $K$ provides a better trade-off: it preserves the main class structure while suppressing redundant components.\par}

	These results support the PCA-like interpretation of the mixed-state prototype. By controlling the rank $K$, the CCPS prototype can filter out low-contribution components that are regarded as noise. Thus, it can focus on the dominant structure of the class distribution. Therefore, the mixed-state prototype is not merely a higher-capacity representation than a pure state, but also provides a controllable low-rank approximation mechanism that improves robustness under incremental learning.

	\section{Conclusion}
	\label{sec:conclusion}
	
	In this paper, we introduced a mixed-state quantum prototype framework for class-incremental learning. Within constrained circuit width, our method matches quantum features with class prototypes to classify. We utilized CCPS ansatz to construct the mixed-state prototypes. It shows higher representational ability than classic single pure state. Furthermore, we designed a prototype-guided replay strategy to achieve incremental learning on quantum hardware. Simulation results indicate that the mixed-state classifier outperforms direct measurement-based quantum classification. And the prototype rank analysis further supported its PCA-like noise-filtering properties. 
	
	For future work, we will explore more hardware-efficient prototype circuits and investigate noise-aware training methodologies on real quantum devices. Additionally, extending this framework to larger-scale incremental scenarios is a promising direction. Accommodating more diverse datasets will further advance quantum models in continual class learning.

	\bibliographystyle{IEEEtran}
	\bibliography{refs}

@inproceedings{rebuffi2017icarl,
	title={icarl: Incremental classifier and representation learning},
	author={Rebuffi, Sylvestre-Alvise and Kolesnikov, Alexander and Sperl, Georg and Lampert, Christoph H},
	booktitle={Proceedings of the IEEE conference on Computer Vision and Pattern Recognition},
	pages={2001--2010},
	year={2017}
}

@article{kirkpatrick2017overcoming,
	title={Overcoming catastrophic forgetting in neural networks},
	author={Kirkpatrick, James and Pascanu, Razvan and Rabinowitz, Neil and Veness, Joel and Desjardins, Guillaume and Rusu, Andrei A and Milan, Kieran and Quan, John and Ramalho, Tiago and Grabska-Barwinska, Agnieszka and others},
	journal={Proceedings of the national academy of sciences},
	volume={114},
	number={13},
	pages={3521--3526},
	year={2017},
	publisher={National Academy of Sciences}
}

@article{li2017learning,
	title={Learning without forgetting},
	author={Li, Zhizhong and Hoiem, Derek},
	journal={IEEE transactions on pattern analysis and machine intelligence},
	volume={40},
	number={12},
	pages={2935--2947},
	year={2017},
	publisher={IEEE}
}

@inproceedings{petit2023fetril,
	title={Fetril: Feature translation for exemplar-free class-incremental learning},
	author={Petit, Gr{\'e}goire and Popescu, Adrian and Schindler, Hugo and Picard, David and Delezoide, Bertrand},
	booktitle={Proceedings of the IEEE/CVF winter conference on applications of computer vision},
	pages={3911--3920},
	year={2023}
}

@article{goswami2023fecam,
	title={Fecam: Exploiting the heterogeneity of class distributions in exemplar-free continual learning},
	author={Goswami, Dipam and Liu, Yuyang and Twardowski, Bart{\l}omiej and Van De Weijer, Joost},
	journal={Advances in Neural Information Processing Systems},
	volume={36},
	pages={6582--6595},
	year={2023}
}

@article{ezzell2023quantum,
	title={Quantum mixed state compiling},
	author={Ezzell, Nic and Ball, Elliott M and Siddiqui, Aliza U and Wilde, Mark M and Sornborger, Andrew T and Coles, Patrick J and Holmes, Zo{\"e}},
	journal={Quantum Science and Technology},
	volume={8},
	number={3},
	pages={035001},
	year={2023},
	publisher={IOP Publishing}
}

@incollection{mccloskey1989catastrophic,
	title={Catastrophic interference in connectionist networks: The sequential learning problem},
	author={McCloskey, Michael and Cohen, Neal J},
	booktitle={Psychology of learning and motivation},
	volume={24},
	pages={109--165},
	year={1989},
	publisher={Elsevier}
}

@article{deng2023novel,
	title={A novel quantum model of mass function for uncertain information fusion},
	author={Deng, Xinyang and Xue, Siyu and Jiang, Wen},
	journal={Information Fusion},
	volume={89},
	pages={619--631},
	year={2023},
	publisher={Elsevier}
}

@article{xiao2022negation,
	title={Negation of the quantum mass function for multisource quantum information fusion with its application to pattern classification},
	author={Xiao, Fuyuan and Pedrycz, Witold},
	journal={IEEE Transactions on Pattern Analysis and Machine Intelligence},
	volume={45},
	number={2},
	pages={2054--2070},
	year={2022},
	publisher={IEEE}
}

@inproceedings{wu2019large,
	title={Large scale incremental learning},
	author={Wu, Yue and Chen, Yinpeng and Wang, Lijuan and Ye, Yuancheng and Liu, Zicheng and Guo, Yandong and Fu, Yun},
	booktitle={Proceedings of the IEEE/CVF conference on computer vision and pattern recognition},
	pages={374--382},
	year={2019}
}

@article{buzzega2020dark,
	title={Dark experience for general continual learning: a strong, simple baseline},
	author={Buzzega, Pietro and Boschini, Matteo and Porrello, Angelo and Abati, Davide and Calderara, Simone},
	journal={Advances in neural information processing systems},
	volume={33},
	pages={15920--15930},
	year={2020}
}

@inproceedings{douillard2020podnet,
	title={Podnet: Pooled outputs distillation for small-tasks incremental learning},
	author={Douillard, Arthur and Cord, Matthieu and Ollion, Charles and Robert, Thomas and Valle, Eduardo},
	booktitle={European conference on computer vision},
	pages={86--102},
	year={2020},
	organization={Springer}
}

@inproceedings{wang2022foster,
	title={Foster: Feature boosting and compression for class-incremental learning},
	author={Wang, Fu-Yun and Zhou, Da-Wei and Ye, Han-Jia and Zhan, De-Chuan},
	booktitle={European conference on computer vision},
	pages={398--414},
	year={2022},
	organization={Springer}
}

@inproceedings{smith2021always,
	title={Always be dreaming: A new approach for data-free class-incremental learning},
	author={Smith, James and Hsu, Yen-Chang and Balloch, Jonathan and Shen, Yilin and Jin, Hongxia and Kira, Zsolt},
	booktitle={Proceedings of the IEEE/CVF international conference on computer vision},
	pages={9374--9384},
	year={2021}
}

@inproceedings{zhu2021prototype,
	title={Prototype augmentation and self-supervision for incremental learning},
	author={Zhu, Fei and Zhang, Xu-Yao and Wang, Chuang and Yin, Fei and Liu, Cheng-Lin},
	booktitle={Proceedings of the IEEE/CVF conference on computer vision and pattern recognition},
	pages={5871--5880},
	year={2021}
}

@inproceedings{zhu2022self,
	title={Self-sustaining representation expansion for non-exemplar class-incremental learning},
	author={Zhu, Kai and Zhai, Wei and Cao, Yang and Luo, Jiebo and Zha, Zheng-Jun},
	booktitle={Proceedings of the IEEE/CVF conference on computer vision and pattern recognition},
	pages={9296--9305},
	year={2022}
}

@inproceedings{belouadah2018deesil,
	title={DeeSIL: Deep-Shallow Incremental Learning.},
	author={Belouadah, Eden and Popescu, Adrian},
	booktitle={Proceedings of the European conference on computer vision (ECCV) workshops},
	pages={0--0},
	year={2018}
}

@inproceedings{hou2019learning,
	title={Learning a unified classifier incrementally via rebalancing},
	author={Hou, Saihui and Pan, Xinyu and Loy, Chen Change and Wang, Zilei and Lin, Dahua},
	booktitle={Proceedings of the IEEE/CVF conference on computer vision and pattern recognition},
	pages={831--839},
	year={2019}
}

@article{zhou2024class,
	title={Class-incremental learning: A survey},
	author={Zhou, Da-Wei and Wang, Qi-Wei and Qi, Zhi-Hong and Ye, Han-Jia and Zhan, De-Chuan and Liu, Ziwei},
	journal={IEEE Transactions on Pattern Analysis and Machine Intelligence},
	volume={46},
	number={12},
	pages={9851--9873},
	year={2024},
	publisher={IEEE}
}

@article{zhang2025few,
	title={Few-shot class-incremental learning for classification and object detection: A survey},
	author={Zhang, Jinghua and Liu, Li and Silv{\'e}n, Olli and Pietik{\"a}inen, Matti and Hu, Dewen},
	journal={IEEE Transactions on Pattern Analysis and Machine Intelligence},
	volume={47},
	number={4},
	pages={2924--2945},
	year={2025},
	publisher={IEEE}
}

@article{van2022three,
	title={Three types of incremental learning},
	author={Van de Ven, Gido M and Tuytelaars, Tinne and Tolias, Andreas S},
	journal={Nature Machine Intelligence},
	volume={4},
	number={12},
	pages={1185--1197},
	year={2022},
	publisher={Nature Publishing Group UK London}
}

@article{shindi2023model,
	title={Model-free quantum gate design and calibration using deep reinforcement learning},
	author={Shindi, Omar and Yu, Qi and Girdhar, Parth and Dong, Daoyi},
	journal={IEEE Transactions on Artificial Intelligence},
	volume={5},
	number={1},
	pages={346--357},
	year={2023},
	publisher={IEEE}
}

@article{dong2019learning,
	title={Learning-based quantum robust control: algorithm, applications, and experiments},
	author={Dong, Daoyi and Xing, Xi and Ma, Hailan and Chen, Chunlin and Liu, Zhixin and Rabitz, Herschel},
	journal={IEEE transactions on cybernetics},
	volume={50},
	number={8},
	pages={3581--3593},
	year={2019},
	publisher={IEEE}
}

@inproceedings{asadi2023prototype,
	title={Prototype-sample relation distillation: towards replay-free continual learning},
	author={Asadi, Nader and Davari, MohammadReza and Mudur, Sudhir and Aljundi, Rahaf and Belilovsky, Eugene},
	booktitle={International conference on machine learning},
	pages={1093--1106},
	year={2023},
	organization={PMLR}
}

@article{masana2022class,
	title={Class-incremental learning: survey and performance evaluation on image classification},
	author={Masana, Marc and Liu, Xialei and Twardowski, Bart{\l}omiej and Menta, Mikel and Bagdanov, Andrew D and Van De Weijer, Joost},
	journal={IEEE Transactions on Pattern Analysis and Machine Intelligence},
	volume={45},
	number={5},
	pages={5513--5533},
	year={2022},
	publisher={IEEE}
}

@inproceedings{gao2022r,
	title={R-dfcil: Relation-guided representation learning for data-free class incremental learning},
	author={Gao, Qiankun and Zhao, Chen and Ghanem, Bernard and Zhang, Jian},
	booktitle={European Conference on Computer Vision},
	pages={423--439},
	year={2022},
	organization={Springer}
}

@article{cerezo2021variational,
	title={Variational quantum algorithms},
	author={Cerezo, Marco and Arrasmith, Andrew and Babbush, Ryan and Benjamin, Simon C and Endo, Suguru and Fujii, Keisuke and McClean, Jarrod R and Mitarai, Kosuke and Yuan, Xiao and Cincio, Lukasz and others},
	journal={Nature Reviews Physics},
	volume={3},
	number={9},
	pages={625--644},
	year={2021},
	publisher={Nature Publishing Group UK London}
}

@article{cerezo2020variational,
	title={Variational quantum fidelity estimation},
	author={Cerezo, Marco and Poremba, Alexander and Cincio, Lukasz and Coles, Patrick J},
	journal={Quantum},
	volume={4},
	pages={248},
	year={2020},
	publisher={Verein zur F{\"o}rderung des Open Access Publizierens in den Quantenwissenschaften}
}

@article{travnivcek2019experimental,
	title={Experimental measurement of the Hilbert-Schmidt distance between two-Qubit states as a means for reducing the complexity of machine learning},
	author={Tr{\'a}vn{\'\i}{\v{c}}ek, Vojt{\v{e}}ch and Bartkiewicz, Karol and {\v{C}}ernoch, Anton{\'\i}n and Lemr, Karel},
	journal={Physical review letters},
	volume={123},
	number={26},
	pages={260501},
	year={2019},
	publisher={APS}
}

@article{yuan2023optimal,
	title={Optimal (controlled) quantum state preparation and improved unitary synthesis by quantum circuits with any number of ancillary qubits},
	author={Yuan, Pei and Zhang, Shengyu},
	journal={Quantum},
	volume={7},
	pages={956},
	year={2023},
	publisher={Verein zur F{\"o}rderung des Open Access Publizierens in den Quantenwissenschaften}
}

@article{du2023problem,
	title={Problem-dependent power of quantum neural networks on multiclass classification},
	author={Du, Yuxuan and Yang, Yibo and Tao, Dacheng and Hsieh, Min-Hsiu},
	journal={Physical Review Letters},
	volume={131},
	number={14},
	pages={140601},
	year={2023},
	publisher={APS}
}

@article{nghiem2021unified,
	title={Unified framework for quantum classification},
	author={Nghiem, Nhat A and Chen, Samuel Yen-Chi and Wei, Tzu-Chieh},
	journal={Physical Review Research},
	volume={3},
	number={3},
	pages={033056},
	year={2021},
	publisher={APS}
}

@article{jiang2022quantum,
	title={Quantum continual learning overcoming catastrophic forgetting},
	author={Jiang, Wenjie and Lu, Zhide and Deng, Dong-Ling},
	journal={Chinese Physics Letters},
	volume={39},
	number={5},
	pages={050303},
	year={2022},
	publisher={Chinese Physical Society and IOP Publishing Ltd}
}

@article{zhang2026experimental,
	title={Experimental demonstration of quantum continual learning with superconducting qubits},
	author={Zhang, Chuanyu and Lu, Zhide and Zhao, Liangtian and Xu, Shibo and Li, Weikang and Wang, Ke and Chen, Jiachen and Wu, Yaozu and Jin, Feitong and Zhu, Xuhao and others},
	journal={npj Quantum Information},
	year={2026},
	publisher={Nature Publishing Group UK London}
}

@article{du2022quantum,
	title={Quantum circuit architecture search for variational quantum algorithms},
	author={Du, Yuxuan and Huang, Tao and You, Shan and Hsieh, Min-Hsiu and Tao, Dacheng},
	journal={npj Quantum Information},
	volume={8},
	number={1},
	pages={62},
	year={2022},
	publisher={Nature Publishing Group UK London}
}

@article{cong2019quantum,
	title={Quantum convolutional neural networks},
	author={Cong, Iris and Choi, Soonwon and Lukin, Mikhail D},
	journal={Nature Physics},
	volume={15},
	number={12},
	pages={1273--1278},
	year={2019},
	publisher={Nature Publishing Group UK London}
}

@article{lloyd2020quantum,
	title={Quantum embeddings for machine learning},
	author={Lloyd, Seth and Schuld, Maria and Ijaz, Aroosa and Izaac, Josh and Killoran, Nathan},
	journal={arXiv preprint arXiv:2001.03622},
	year={2020}
}

@article{maccormack2022branching,
	title={Branching quantum convolutional neural networks},
	author={MacCormack, Ian and Delaney, Conor and Galda, Alexey and Aggarwal, Nidhi and Narang, Prineha},
	journal={Physical Review Research},
	volume={4},
	number={1},
	pages={013117},
	year={2022},
	publisher={APS}
}

@article{schuld2020circuit,
	title={Circuit-centric quantum classifiers},
	author={Schuld, Maria and Bocharov, Alex and Svore, Krysta M and Wiebe, Nathan},
	journal={Physical Review A},
	volume={101},
	number={3},
	pages={032308},
	year={2020},
	publisher={APS}
}

@article{grant2018hierarchical,
	title={Hierarchical quantum classifiers},
	author={Grant, Edward and Benedetti, Marcello and Cao, Shuxiang and Hallam, Andrew and Lockhart, Joshua and Stojevic, Vid and Green, Andrew G and Severini, Simone},
	journal={npj Quantum Information},
	volume={4},
	number={1},
	pages={65},
	year={2018},
	publisher={Nature Publishing Group UK London}
}

@article{stein2022quclassi,
	title={Quclassi: A hybrid deep neural network architecture based on quantum state fidelity},
	author={Stein, Samuel A and Baheri, Betis and Chen, Daniel and Mao, Ying and Guan, Qiang and Li, Ang and Xu, Shuai and Ding, Caiwen},
	journal={Proceedings of Machine Learning and Systems},
	volume={4},
	pages={251--264},
	year={2022}
}

@inproceedings{mensink2012metric,
	title={Metric learning for large scale image classification: Generalizing to new classes at near-zero cost},
	author={Mensink, Thomas and Verbeek, Jakob and Perronnin, Florent and Csurka, Gabriela},
	booktitle={European conference on computer vision},
	pages={488--501},
	year={2012},
	organization={Springer}
}

@inproceedings{yu2020semantic,
	title={Semantic drift compensation for class-incremental learning},
	author={Yu, Lu and Twardowski, Bart{\l}omiej and Liu, Xialei and Herranz, Luis and Wang, Kai and Cheng, Yongmei and Jui, Shangling and Van de Weijer, Joost},
	booktitle={2020 IEEE/CVF Conference on Computer Vision and Pattern Recognition (CVPR)},
	pages={6980--6989},
	year={2020},
	organization={IEEE}
}

@inproceedings{tao2020few,
	title={Few-shot class-incremental learning},
	author={Tao, Xiaoyu and Hong, Xiaopeng and Chang, Xinyuan and Dong, Songlin and Wei, Xing and Gong, Yihong},
	booktitle={2020 IEEE/CVF Conference on Computer Vision and Pattern Recognition (CVPR)},
	pages={12180--12189},
	year={2020},
	organization={IEEE}
}

@inproceedings{liu2020more,
	title={More classifiers, less forgetting: A generic multi-classifier paradigm for incremental learning},
	author={Liu, Yu and Parisot, Sarah and Slabaugh, Gregory and Jia, Xu and Leonardis, Ales and Tuytelaars, Tinne},
	booktitle={European Conference on Computer Vision},
	pages={699--716},
	year={2020},
	organization={Springer}
}

@article{shi2026qsyncfold,
	title={QSyncFold: quantum neural network for multidimensional sync-discovery in protein folding},
	author={Shi, Jinjing and Du, Peng and Zeng, Wenwu and Wang, Wenxuan and Peng, Shaoliang and Li, Xuelong},
	journal={Briefings in Bioinformatics},
	volume={27},
	number={3},
	pages={bbag234},
	year={2026},
	publisher={Oxford University Press}
}

@article{shi2025quantum,
	title={Quantum option profit and loss encoding for risk assessment},
	author={Shi, Jinjing and Yuan, Bingjie and Lu, Yuhu and Zhang, Shichao and Li, Xuelong},
	journal={IEEE Transactions on Emerging Topics in Computational Intelligence},
	year={2025},
	publisher={IEEE}
}

@article{shi2024qsan,
	title={QSAN: A near-term achievable quantum self-attention network},
	author={Shi, Jinjing and Zhao, Ren-Xin and Wang, Wenxuan and Zhang, Shichao and Li, Xuelong},
	journal={IEEE Transactions on Neural Networks and Learning Systems},
	volume={36},
	number={8},
	pages={13995--14008},
	year={2024},
	publisher={IEEE}
}

@article{wu2026feature,
	title={Feature Entanglement-based Quantum Multimodal Fusion Neural Network},
	author={Wu, Yu and Zhou, Qianli and Geng, Jie and Deng, Xinyang and Jiang, Wen},
	journal={arXiv preprint arXiv:2601.07856},
	year={2026}
}

@article{zhao2026multidomain,
	title={Multidomain Evolutionary Optimization on Adversarial Link Perturbation in Imbalanced-Size Complex Systems},
	author={Zhao, Jie and Cheong, Kang Hao},
	journal={IEEE Transactions on Systems, Man, and Cybernetics: Systems},
	year={2026},
	publisher={IEEE}
}

@article{zhan2026ternary,
	title={Ternary coding of maximum Deng entropy},
	author={Zhan, Tianxiang and He, Yuanpeng and Deng, Yong},
	journal={Fuzzy Sets and Systems},
	pages={109913},
	year={2026},
	publisher={Elsevier}
}

@article{su2026dependence,
	title={A dependence assessment method based on quantum model of mass function in human reliability analysis},
	author={Su, Xiaoyan and Huang, Xuying and Pan, Xiaolei and Meng, Debiao},
	journal={Expert Systems with Applications},
	volume={299},
	pages={129992},
	year={2026},
	publisher={Elsevier}
}

@article{yang2025multimodal,
	title={Multimodal remote sensing of thunderstorm charge motion: A radar echo and electric field fusion approach},
	author={Yang, Xu and Xing, Hongyan and Zhu, Fa and Chen, Yong and Camacho, David and Dong, Xudong and Pedrycz, Witold},
	journal={IEEE Transactions on Geoscience and Remote Sensing},
	volume={63},
	pages={1--13},
	year={2025},
	publisher={IEEE}
}

@article{huang2026individual,
	title={Individual Linguistic Granular Computing: A Granulation--Degranulation-Based Approach},
	author={Huang, Ting and Zhang, Qiang and Pedrycz, Witold and Yang, Shanlin},
	journal={IEEE Transactions on Cybernetics},
	year={2026},
	publisher={IEEE}
}

@article{xiao2026adaptive,
	title={An Adaptive Quantum Circuit of Dempster's Rule of Combination for Uncertain Pattern Classification},
	author={Xiao, Fuyuan and Zhou, Yu and Pedrycz, Witold},
	journal={Advances in Neural Information Processing Systems},
	volume={38},
	pages={17848--17881},
	year={2026}
}

@article{xu2024quantum,
	title={Quantum classifiers with a trainable kernel},
	author={Xu, Li and Zhang, Xiao-yu and Li, Ming and Shen, Shu-qian},
	journal={Physical Review Applied},
	volume={21},
	number={5},
	pages={054056},
	year={2024},
	publisher={APS}
}

@article{li2025efficient,
	title={An efficient quantum proactive incremental learning algorithm},
	author={Li, Lingxiao and Li, Jing and Song, Yanqi and Qin, Sujuan and Wen, Qiaoyan and Gao, Fei},
	journal={Science China Physics, Mechanics \& Astronomy},
	volume={68},
	number={1},
	pages={210313},
	year={2025},
	publisher={Springer}
}

@article{huang2026incomplete,
	title={Incomplete data classification via distribution alignment with evidence combination},
	author={Huang, Linqing and Fan, Jinfu and Wang, Shilin and Liu, Gongshen and Liew, Alan Wee-Chung},
	journal={Machine Intelligence Research},
	volume={23},
	number={3},
	pages={593--615},
	year={2026},
	publisher={Springer}
}

@article{huang2026dual,
	title={Dual Joint Covariance Alignment Method for Incomplete Data Classification},
	author={Huang, Linqing and Liu, Gongshen},
	journal={IEEE Transactions on Neural Networks and Learning Systems},
	year={2026},
	publisher={IEEE}
}
	
	\vfill

\end{document}